\documentclass[sigconf]{acmart}

\usepackage{subfig}
\usepackage{enumitem}
\usepackage{graphicx}  
\usepackage{xcolor}    
\usepackage{rotating}
\usepackage{multirow}
\usepackage{makecell}
\usepackage{pifont}
\usepackage{hyperref}

\definecolor{lightgray}{gray}{0.9}
\newcommand{\cmark}{\textcolor{green}{\ding{51}}}

\AtBeginDocument{%
  }

\copyrightyear{2025}
\acmYear{2025}
\setcopyright{cc}
\setcctype{by}
\acmConference[CHI '25]{CHI Conference on Human Factors in Computing Systems}{April 26-May 1, 2025}{Yokohama, Japan}
\acmBooktitle{CHI Conference on Human Factors in Computing Systems (CHI '25), April 26-May 1, 2025, Yokohama, Japan}\acmDOI{10.1145/3706598.3713513}
\acmISBN{979-8-4007-1394-1/25/04}

\thanks{© {Owner/Author | ACM} {2025}. This is the author's version of the work. It is posted here for your personal use. Not for redistribution. The definitive Version of Record was published in CHI Conference on Human Factors in Computing Systems (CHI '25), https://doi.org/10.1145/3706598.3713513.}

\begin{document}

\title{From Passive Watching to Active Learning: Empowering Proactive Participation in Digital Classrooms with AI Video Assistant}


\author{Anna Bodonhelyi}
\orcid{0009-0001-8941-3893}
\affiliation{%
  \institution{Chair of Human-Centered Technologies for Learning, TUM}
  \city{Munich}
  %\state{Bayern}
  \country{Germany}}
\email{anna.bodonhelyi@tum.de}

\author{Enkeleda Thaqi}
\affiliation{%
  \institution{Chair of Human-Centered Technologies for Learning, TUM}
  \city{Munich}
  \country{Germany}}
\email{enkeleda.thaqi@tum.de}

\author{Süleyman Özdel}
\affiliation{%
  \institution{Chair of Human-Centered Technologies for Learning, TUM}
  \city{Munich}
  \country{Germany}}
  \email{ozdelsuleyman@tum.de}

\author{Efe Bozkir}
\affiliation{%
 \institution{Chair of Human-Centered Technologies for Learning, TUM}
 \city{Munich}
 \country{Germany}}
 \email{efe.bozkir@tum.de}

\author{Enkelejda Kasneci}
\affiliation{%
  \institution{Chair of Human-Centered Technologies for Learning, TUM}
  \city{Munich}
  \country{Germany}}
\email{enkelejda.kasneci@tum.de}

\renewcommand{\shortauthors}{Anna Bodonhelyi, Enekelda Thaqi, Süleyman Özdel, Efe Bozkir, Enkelejda Kasneci}


\begin{abstract}
In online education, innovative tools are crucial for enhancing learning outcomes. SAM (Study with AI Mentor) is an advanced platform that integrates educational videos with a context-aware chat interface powered by large language models. SAM encourages students to ask questions and explore unclear concepts in real time, offering personalized, context-specific assistance, including explanations of formulas, slides, and images. We evaluated SAM in two studies: one with 25 university students and another with 80 crowdsourced participants, using pre- and post-knowledge tests to compare a group using SAM and a control group. The results demonstrated that SAM users achieved greater knowledge gains specifically for younger learners and individuals in flexible working environments, such as students, supported by a 97.6\% accuracy rate in the chatbot's responses. Participants also provided positive feedback on SAM’s usability and effectiveness. SAM’s proactive approach to learning not only enhances learning outcomes but also empowers students to take full ownership of their educational experience, representing a promising future direction for online learning tools.
\end{abstract}





\maketitle

\section{Introduction}
Chatbots have increasingly become valuable tools in educational settings, offering significant enhancements to the learning experience and outcomes~\cite{hobert2019say, kuhail2023interacting}. Research consistently shows that these artificial intelligence (AI)-driven tools can increase student engagement, motivation, and knowledge retention by providing personalized support and immediate feedback~\cite{pappagallo2024chatbots, ghayoomi2023applications, tapalova2022artificial, benotti2017tool, kuhail2023interacting, wu2024ai}. Chatbots have been effectively used to deliver course materials, assist with assignments~\cite{sinha2020educational}, and foster interactive learning environments~\cite{hobert2019say, kuhail2023interacting}, leading to improved academic performances~\cite{pappagallo2024chatbots} across various disciplines. However, these tools are not without their limitations. Common drawbacks include the potential for contextually irrelevant or inaccurate responses~\cite{goodman2023accuracy}, limited ability to handle complex queries~\cite{zhang2024chatbots}, and reliance on pre-programmed content~\cite{ranoliya2017chatbot}, which can restrict the depth and adaptability of interactions.

Building on the growing influence of AI in education~\cite{kasneci2023chatgpt}, our tool, SAM (Study with AI Mentor), significantly enhances the learning experience. SAM is an innovative and context-aware platform designed to assist learners in grasping complex topics through interactive, AI-driven mentoring. SAM is built to function as an intelligent video-watching platform where users can watch educational videos, particularly lectures, and engage with an AI mentor to ask questions related to the content. This real-time assistance aims to deepen students' comprehension by providing immediate, context-relevant feedback, thereby simulating an interactive learning environment. Furthermore, SAM builds on existing literature that demonstrates how asking questions while studying leads to increased engagement and, consequently, improved learning outcomes~\cite{zeegers2019s, buhler2023automated, sedova2019those}. By fostering this proactive behavior, SAM contributes to a more effective and engaging learning process.

The core functionality of SAM revolves around its ability to integrate multiple sources of information. Users can provide a link to the lecture video and optionally upload the associated slides, allowing SAM to rely on both the video transcript and the slides to deliver more accurate and informative responses. As the video progresses, users can interact with the AI mentor by asking questions directly related to the content they view. We designed SAM to be responsive, recognizing specific references to slides or concepts and generating detailed explanations. Additionally, SAM employs LaTeX for rendering complex mathematical formulas, ensuring that all visual elements are presented clearly and accurately, which is essential for subjects that rely heavily on precise notation. Upon completing the video, users could download a PDF summarizing their questions and the AI’s responses, providing a record of their learning process.

For the purpose of the user study, we tailored SAM to support a crowdsourced user study where participants engaged with a pre-selected lecture video. To this end, the platform featured a simplified interface, allowing participants to start the already uploaded video, ensuring uniform interaction with the content and features across the study. In this study, we focused on measuring the knowledge gain to provide insights into how a dynamic and real-time chatbot assistant like SAM can enhance learning outcomes. By examining the impact of SAM on participants' comprehension and retention, we aimed to better understand the effectiveness of integrating AI-driven, real-time assistance into educational environments.

While chatbots have significantly advanced education, critical gaps remain, particularly in fields requiring complex, context-specific explanations that SAM specifically addresses. SAM represents a new way of learning, where learners take full ownership of their educational journey. By encouraging proactive behavior, SAM not only enhances engagement—a factor already shown to be effective in previous studies~\cite{zeegers2019s, buhler2023automated, sedova2019those}—but also fosters a more interactive and responsive learning environment. Through our user study with SAM, we provide valuable insights into future online learning techniques, offering guidance on how to design tools that trigger and sustain interactive learning experiences. The main contributions of this work are as follows:

\begin{itemize}
    \item Development of SAM, an AI-driven educational tool that enhances learning by integrating real-time question-answering with video content, supporting both text and image-based queries, and accurately rendering complex mathematical formulas.
    \item Comprehensive evaluation of SAM's impact on knowledge gain and user satisfaction through crowdsourcing, highlighting its effectiveness and positive effect. The version, which was used to this study is available at \url{https://gitlab.lrz.de/hctl/sam}.
    \item Identification of key demographic subgroups, such as students and younger users, that demonstrated the highest knowledge gains when using SAM, providing valuable insights for tailoring future educational tools.
\end{itemize}

\section{Related Work}
AI has been significantly transforming education by enabling personalized learning experiences, enhancing student engagement, and optimizing administrative tasks~\cite{pratama2023revolutionizing, benotti2017tool}. AI technologies, such as machine learning and natural language processing, allow for tailored educational approaches that cater to individual learning styles and paces, potentially leading to improved student outcomes~\cite{harry2023role}. Recent studies further highlight the potential of tools like ChatGPT to support education by enabling widespread usage, though careful consideration of student backgrounds and concerns is essential for effective implementation~\cite{stohr2024perceptions}. This section highlights key advancements relevant to SAM, examining online learning tools, chatbots, virtual tutors, their overall impact on learning outcomes, and the relationship between student engagement and question-asking in classrooms.

\subsection{Overview of Online Learning Tools}
Online learning tools have revolutionized education, providing flexible and accessible options for students and educators worldwide. These digital platforms and software applications facilitate remote learning, content creation, student engagement, and assessment in virtual environments~\cite{al2020evaluating, martin2018engagement}. They also have a crucial role in modern education, especially in the context of engineering disciplines~\cite{sivapalan2016engineering}, science education~\cite{rutten2012learning}, language learning~\cite{gonzalez2020collaborative}, and overall digital transformation of education~\cite{beaumont2018google}. These tools, ranging from digital collaboration platforms like Zoom~\cite{minhas2021exploring, gunawan2021application}, Microsoft Teams~\cite{rojabi2020exploring}, and Google Classroom~\cite{beaumont2018google} to interactive video-based learning and simulation tools~\cite{gordillo2022comparing}, have been shown to enhance student engagement, motivation, and learning outcomes in various educational settings~\cite{vermeulen2024promoting}.
By incorporating features like learner-centered approaches, fast feedback mechanisms, and modern technologies, online learning tools not only improve the quality of education but also make the learning process more efficient for students~\cite{pokotylo2023some}. Additionally, the use of various approaches and tools like gamification, simulation, and virtual reality have been experimented with to enhance knowledge dissemination and learning experiences~\cite{sinha2024tools}. As technology continues to evolve, further research is needed to explore the full potential of these tools in diverse educational contexts and student populations.

One of the most notable trends in online education is the rise of Massive Open Online Courses (MOOCs), which offer high-quality courses for mass delivery through the Internet~\cite{dabbagh2016massive, ferreira2016massive}. Initially coined in 2008~\cite{fini2009technological, liyanagunawardena2015massive}, MOOCs have gained popularity globally, attracting learners interested in experiencing top-tier university teaching~\cite{liyanagunawardena2015massive}. These courses have disrupted conventional educational models by offering free, accessible, and diverse learning opportunities to a broad audience, challenging the economic and pedagogical paradigms of traditional on-campus universities~\cite{ferreira2016massive}. With over 11,000 MOOCs available worldwide as of 2022~\cite{gupta2022massive}, these courses have revolutionized education delivery, reaching thousands of students and prompting discussions on their impact on international communication and learning models. Platforms such as Coursera~\cite{ayoub2020contribution}, edX~\cite{sanchez2016could}, and Udacity~\cite{anyatasia2020evaluation} have expanded access to education by offering courses from top universities and institutions to a global audience~\cite{tsironis2016comparative}, often for free or at a low cost. MOOCs cover many subjects and provide learners with flexibility and accessibility, accommodating diverse schedules and learning paces. These platforms also incorporate interactive elements such as peer assessments, discussion forums, and real-time feedback to enhance the learning experience.

In addition to MOOCs, video-based learning platforms have also gained popularity by offering benefits such as enhanced student engagement and motivation, flexibility and accessibility for learners, and support for teacher professional development~\cite{sablic2021video}. Additionally, using video-based student support platforms, like ORBITS~\cite{shehata2023enhancing} predictive engine, has positively enhanced student learning experiences through AI techniques. Platforms such as Panopto~\cite{getenet2022students} and Kaltura~\cite{govender2022first} provide comprehensive video hosting and streaming services, which include features like interactive quizzes and analytics to track learner engagement and progress. These tools are designed to cater to various learning styles by incorporating visual and auditory elements, making complex subjects more accessible and engaging for students~\cite{sheridan2024panopto}. While these platforms offer valuable features, SAM goes a step further by providing real-time, personalized support during learning sessions, enabling students to ask questions and receive immediate, context-specific feedback. This level of interactivity helps bridge knowledge gaps more effectively than traditional video-based platforms, enhancing the overall learning experience.

\subsection{Chatbots and Virtual Tutors in Learning}
Chatbots are increasingly recognized for their transformative educational potential, leveraging AI to enhance learning experiences. They offer personalized and continuous support, which can significantly improve student engagement and motivation, particularly in e-learning environments where traditional interaction may be limited~\cite{pappagallo2024chatbots, ghayoomi2023applications, tapalova2022artificial, benotti2017tool, kuhail2023interacting}. They can enhance learning by providing course materials~\cite{cunningham2019visualizing}, assignments and practice questions~\cite{sinha2020educational}, while also engaging students individually~\cite{hobert2019say} or facilitating group learning activities~\cite{tegos2014configurable, kuhail2023interacting}. Chatbots usually engage learners by utilizing text, speech, visuals, touch, and gestures to support them in different educational activities~\cite{kuhail2023interacting}. Previous research indicates that chatbots can foster higher-order cognitive skills and reduce dropout rates by adapting to individual student needs~\cite{pappagallo2024chatbots}. A recent study~\cite{kuhail2023interacting} indicates that many previous works have predominantly utilized chatbots in the fields of computer science and language education. In these contexts, chatbots can assume various roles to enhance learning, most commonly serving as teaching~\cite{coronado2018cognitive, song2017interacting} or peer agents~\cite{lee2020using, da2019ibm}, but also functioning as improving~\cite{law2020curiosity} or motivational agents~\cite{schouten2017digital, kuhail2023interacting, wu2024ai}. Teaching agents employ a variety of approaches to enhance learning. These methods include recommending personalized tutorials, initiating discussions based on educational videos or past learning, presenting scenario-based exercises, and providing assessments through multiple-choice questions. On the other hand, peer agent chatbots primarily support students by offering on-demand assistance, including answering specific questions, clarifying terms or concepts, providing guidance on using certain technologies, and facilitating group discussions~\cite{kuhail2023interacting}. 

Several chatbot implementations highlight these roles, showcasing how chatbots can effectively address diverse educational needs across various subjects. For instance, the FAQ chatbot~\cite{han2022faq} assists students in MOOC classes by answering administrative questions, leveraging a pre-trained large language model (LLM) based on information from a FAQ website. Similar to the FAQ chatbot~\cite{han2022faq}, Cipherbot~\cite{salminen2024using} was developed to answer students' questions based on educational materials that were uploaded by educators to minimize hallucinations. Powered by GPT-3.5 Turbo, Cipherbot achieved an 87.0\% accuracy rate in its evaluation study. Both tools are designed to respond to course- or material-specific questions; however, while both utilize externally validated sources, their potential for a close integration with a video player remains to be investigated. To bridge this gap, Sara, the lecturer~\cite{winkler2020sara}, offers scaffolding support during lecture viewing by prompting students with questions and exercises at predefined points within the video. Sara guides learners through these tasks, providing scaffolds until they can solve the posed problems independently. This tool prioritizes engaging learners through inquiry and active questioning rather than directly answering user queries~\cite{winkler2020sara}. Unlike these tools, we designed our framework to provide real-time support during lecture viewing by responding to any student question with high-quality, accurate answers related to the lecture content, focusing specifically on seamless integration with the lecture material to enhance understanding.

On the other hand, there are notable challenges associated with using chatbots. Their integration also raises ethical concerns, including issues of privacy, data manipulation, and the potential for academic dishonesty, particularly with tools like ChatGPT~\cite{OpenAI2023, yang2024chatgpt, lo2024influence}. Research indicates that user privacy is a significant issue, with self-disclosure being a primary concern due to the sensitive data shared during interactions with chatbots~\cite{ischen2020privacy, gumusel2024literature}. Furthermore, the misuse of chatbots can lead to the dissemination of incorrect or harmful safety-related information, emphasizing the need for expert verification and ethical safeguards~\cite{zhou2023ethical, lo2024influence}. Additionally, as LLMs like GPT-4 evolve, challenges in accurately responding to specific user intents emphasize the need for improvements in intent recognition to enhance user satisfaction and ensure more reliable interactions~\cite{bodonhelyi2024user}. Overall, while chatbots offer substantial benefits, these concerns highlight the necessity for responsible usage and regulatory frameworks to mitigate risks associated with privacy, ethics, and information accuracy.

While chatbots can facilitate language learning and stimulate curiosity through project-driven tasks, their reliance on extensive knowledge bases poses challenges in development and implementation~\cite{ghayoomi2023applications}, often leading to difficulties in handling complex messages effectively~\cite{zhang2024chatbots}. This dependency can lead to instances of chatbot fallibility, where the information provided may be inaccurate or misleading~\cite{goodman2023accuracy}. Such reliability issues can undermine the learning experience, necessitating careful oversight and regular updates to ensure that chatbots remain accurate and trustworthy educational tools~\cite{roldan2024reliability}. Educators must be aware of these limitations and supplement chatbot interactions with human guidance to maintain the quality of instruction. To address these challenges, SAM focuses on increasing answer accuracy by providing context-aware responses. By integrating real-time feedback with relevant contextual information, SAM aims to minimize errors and enhance the reliability of its educational support.

\subsection{Engagement Through Questions}

\begin{figure*}[t]
    \centering
    \includegraphics[height=6.5cm, keepaspectratio]{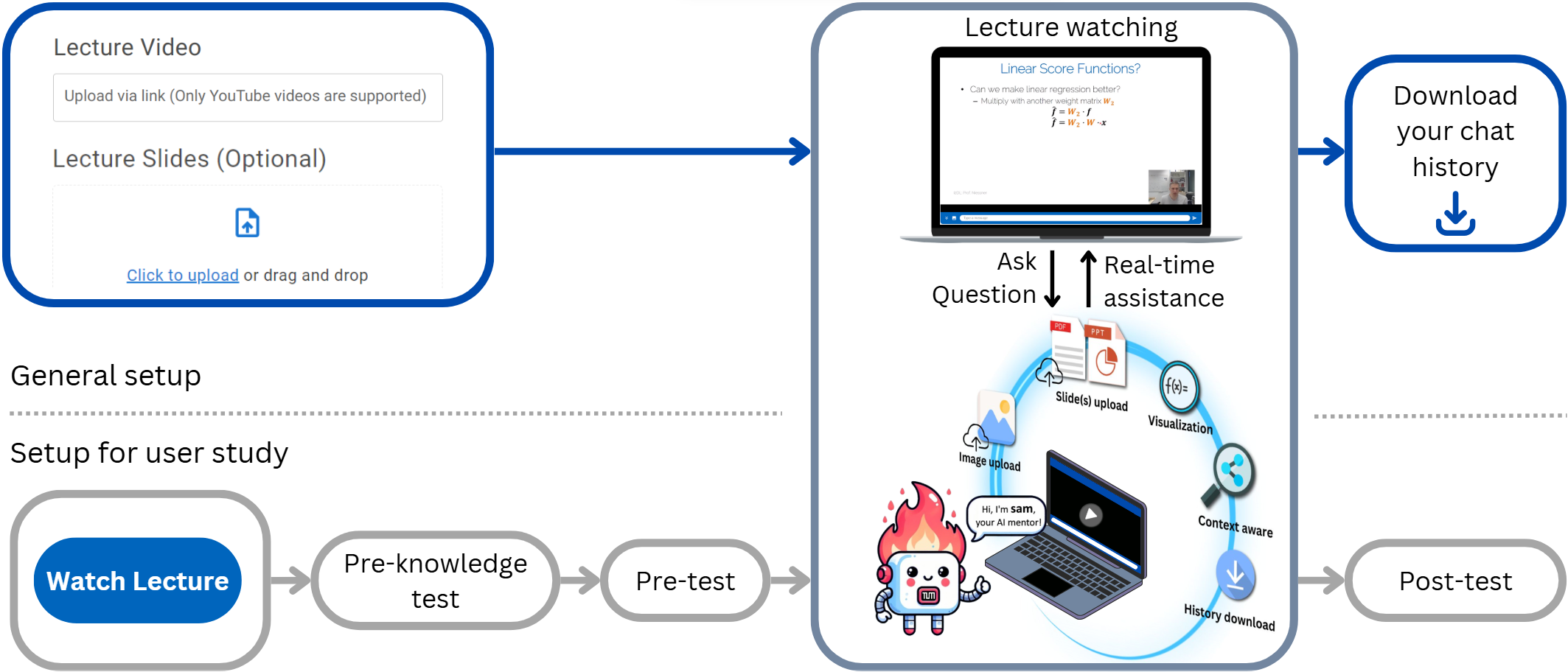}
    \caption{Structure of SAM, with modification for the user study.}
    \label{fig:concept}
\end{figure*}

Learner engagement, which refers to the investment of students' cognitive and emotional effort in completing a learning task~\cite{lam2012understanding}, has been shown to positively influence key educational outcomes such as academic performance, persistence, satisfaction, and a sense of community~\cite{wang2014staying, halverson2019learner}. In online settings, engagement often encompasses not only traditional indicators like contributing to discussions and asking questions~\cite{zeegers2019s, buhler2023automated}, but also involves consistent participation in virtual activities such as live chats~\cite{hew2023using}, quizzes~\cite{raes2020learning}, and collaborative tasks~\cite{kim2015connecting}. In online learning, where distractions can be more common~\cite{buhler2024task}, engagement becomes crucial to ensure students remain focused, driving better learning outcomes. Effective online engagement also involves leveraging digital tools that promote interaction and personalized learning experiences. AI-driven approaches, such as automated classroom observation protocols, can provide teachers with specific feedback on their practices, streamlining the assessment process while ensuring consistent and valuable insights~\cite{hou2024automated}.

Proactive behaviors, such as students actively asking questions, serve as key indicators of engagement in online learning environments~\cite{zeegers2019s, buhler2023automated, sedova2019those}. When students take the initiative to ask about course content, it reflects a high level of cognitive engagement, pointing out that they are not passively absorbing information but critically processing it. Questioning is an essential component of active learning, as it demonstrates curiosity and a desire to deepen understanding, driving students to connect new knowledge with prior learning~\cite{boheim2020student, boheim2020exploring}.

Taking all into account, SAM aligns with and extends theoretical perspectives on active and interactive learning, particularly as articulated in the ICAP framework~\cite{chi2014icap}. This framework posits that learning outcomes improve progressively from passive to active, constructive, and interactive engagement levels. By integrating real-time and context-aware interactions into video-based learning, SAM elevates the traditional passive consumption of educational content to an interactive learning experience. Unlike previous approaches that rely on pre-scripted prompts or rigidly structured content, SAM's ability to respond dynamically to user queries fosters higher cognitive engagement and deepens comprehension~\cite{zeegers2019s, buhler2023automated, sedova2019those}. This interactive paradigm bridges a critical gap in the literature by demonstrating how AI-driven pedagogical tools can operationalize theoretical constricts of interactive learning in a scalable, real-world application.

\section{SAM Design and Implementation}
SAM is an innovative tool designed to enhance educational experiences by integrating AI-driven assistance with video-based learning. SAM operates as an advanced video-watching platform, enabling users to view YouTube lectures while interacting with an AI mentor to pose questions related to the content (\autoref{fig:concept}). The integrated AI mentor is powered by GPT-4o~\cite{gpt4o}, a highly advanced language model that offers sophisticated natural language processing capabilities. The AI mentor not only supports the instant clarification of doubts but also promotes a deeper understanding of complex topics by delivering explanations tailored to learners' needs. This dynamic and personalized learning experience ultimately helps students grasp the subject matter more effectively, enhancing their overall educational journey.

SAM actively fosters interactive learning, aligning with the principles highlighted in the ICAP framework~\cite{chi2014icap}, which identifies interactivity as the most effective level for enhancing learning compared to constructive, active, or passive approaches, thereby ensuring an optimal learning experience. The user interface design of SAM prioritizes simplicity and aesthetic appeal, keeping in mind previous findings that users tend to favor interfaces that are visually attractive and easy to navigate over overly complex designs~\cite{senevirathne2021impact}. Research suggests a strong positive relationship between user satisfaction and key factors such as learnability, visibility, and interactivity~\cite{senevirathne2021impact}. In SAM, the chat window is strategically placed at the bottom of the interface, differing from the conventional right-side placement commonly seen in streaming platforms. This bottom placement allows for a resizable chat window that partially overlaps with the video. However, to enhance user focus, the video automatically pauses when users begin typing, enabling them to engage more effectively with the AI tutor and view detailed responses, including formulas, without distraction.

\begin{figure*}[t]
	\centering
	{\includegraphics[height=5.2cm, keepaspectratio]{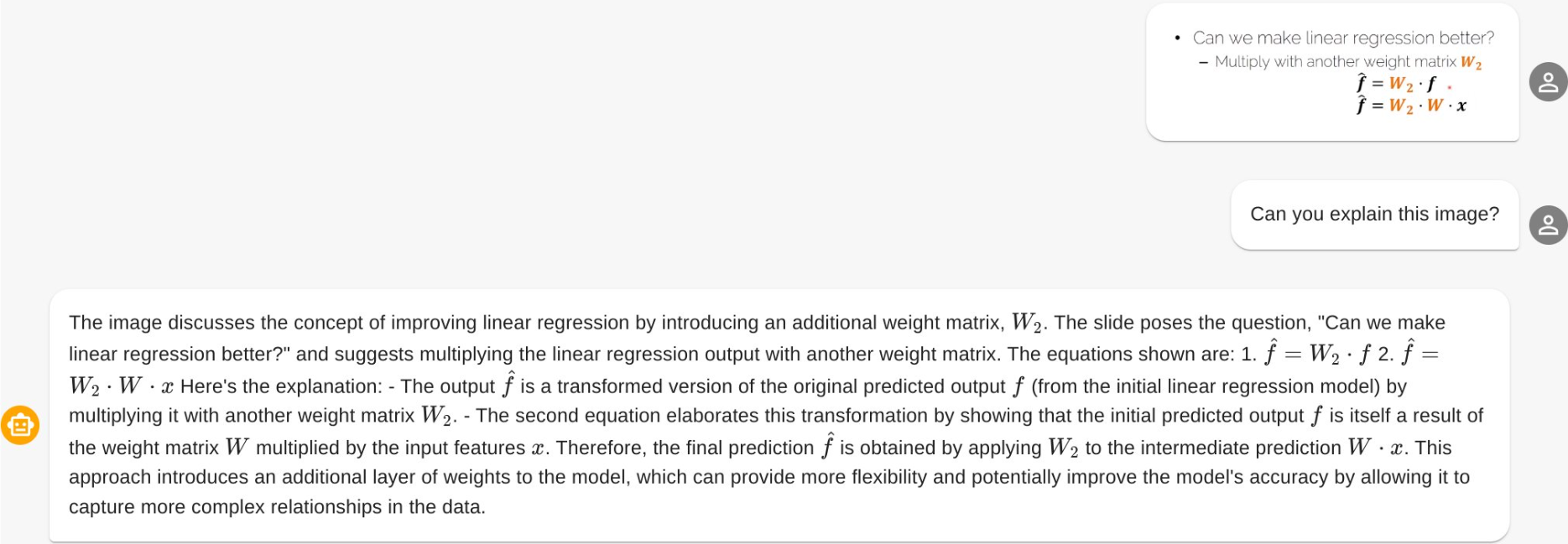}}
	\caption{Example of SAM in action: user submission of an image and formula visualization.}
    \label{fig:img_form}
\end{figure*}

The user experience with SAM begins when a lecture link from YouTube is provided. Users may also upload corresponding slides with the possibility to also upload previous slides or relevant related materials, which enriches the AI mentor's ability to deliver accurate and context-aware responses by leveraging both the video transcript and the additional slide information. To achieve highly accurate answers and only answer lecture-related questions, the following prompt was used:\\

\noindent
\colorbox{lightgray}{%
    \parbox{\dimexpr\linewidth-2\fboxsep\relax}{%
        \texttt{You are roleplaying as an assistant teacher helping students understand their lecture content. Answer their questions based on the video, video transcripts, and slides. If a question is about a lecture unrelated topic, respond with ``Please focus on the lecture material.'' If you don't know the answer, just say that you don't know. Do not attempt to fabricate an answer.}
    }%
}\\

During video playback, users can engage the AI mentor by entering queries in a designated text field and the AI mentor formulates responses to the user inquiries. When a user submits a message, the lecture slides, transcript, the 30 seconds of the transcript before and after the current timestamp, and the chat history are sent to the LLM. The video itself is not forwarded. The timely relevant parts of the transcript are also provided, offering additional context to the embedded language model. Additionally, SAM allows users to upload images (e.g., screenshots from the slides), further aiding the LLM in providing precise and contextually relevant answers (\autoref{fig:img_form}). SAM is designed with features to enhance the learning process, including the automatic transmission of the current slide whenever the word ``slide'' appears in a user's message. For instance, if a user asks, ``Explain this slide!'' the slide is sent as an image along with the query to the AI mentor, significantly increasing the accuracy and relevance of the response. Prior to the studies, we specifically tested this feature and found that including the slide improved answer accuracy, as the LLM often struggled to pinpoint the exact content of the current slide without this contextual aid. Moreover, SAM supports LaTeX-type formatting for visualizing mathematical formulas, ensuring that complex expressions are displayed clearly, facilitating better comprehension (\autoref{fig:img_form}). Upon completion of the session, users can download a PDF containing the questions posed and the AI mentor’s responses, serving as a valuable study resource. 

\begin{figure*}[t]
	\centering
	{\includegraphics[height=4cm, keepaspectratio]{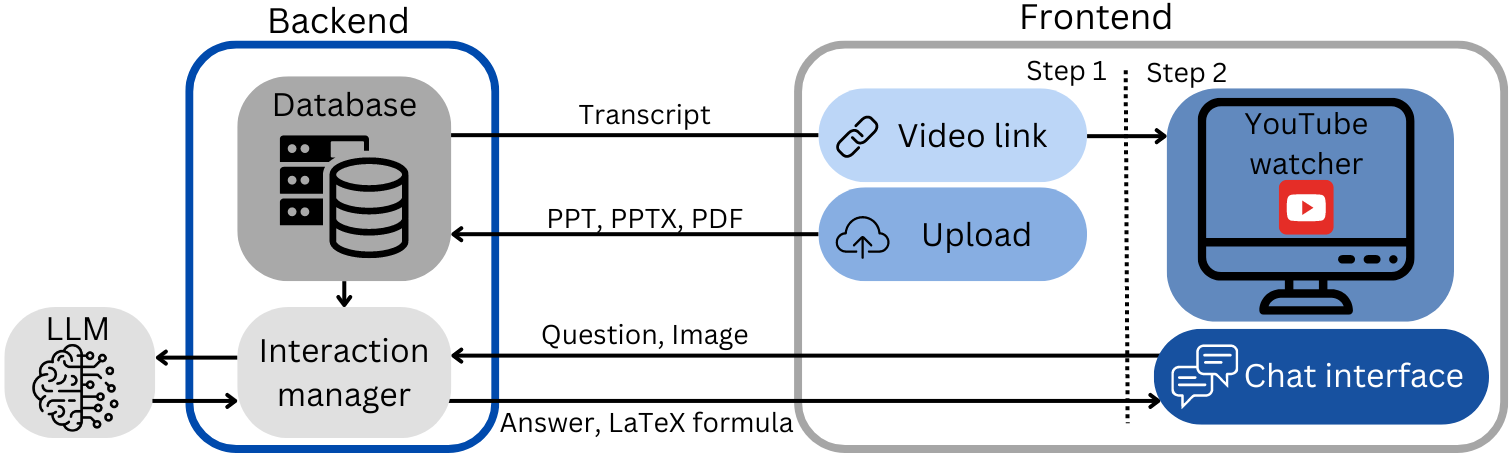}}
	\caption{Implemented modules and their connections in the web application of SAM.}
    \label{fig:web}
\end{figure*}

To support the smooth operation of SAM, the platform's architecture (\autoref{fig:web}) was designed with distinct frontend and backend components, each hosted on separate servers. The frontend, deployed on a dedicated server, is responsible for the user interface and interaction. The backend, hosted on Amazon's Web Services (AWS) server~\cite{aws}, handles the core processing and data management tasks. AWS~\cite{aws} was chosen for its robust infrastructure, scalability, and reliability, allowing SAM to efficiently manage multiple user sessions simultaneously. Each user's data is stored under a specific but anonymous ID, enabling the platform to securely save their interactions and responses, which are also needed for the resulting downloadable PDF. These records are then compiled into a resulting PDF and JSON files for further analysis. The coding of SAM leverages Node Package Manager for managing dependencies, with the core files written in TSX (TypeScript and JSX), a powerful combination that enhances both the type safety and flexibility of the development process. This technical setup ensures that SAM remains a robust, scalable, and user-friendly tool for enhancing educational experiences.

\section{Methodology}
\begin{figure*}[t]
	\centering
	{\includegraphics[height=3.7cm, keepaspectratio]{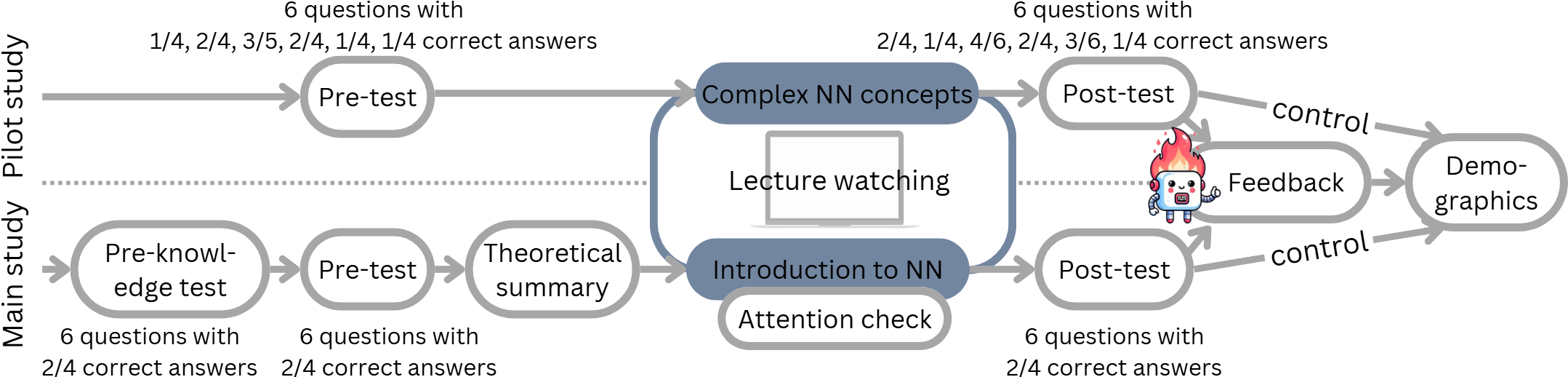}}
	\caption{Representation of the structural elements in the pilot and main studies.}
    \label{fig:p_m}
\end{figure*}

To explore how proactive behavior can be encouraged in online learning environments, we designed SAM to provide learners with personalized, real-time support while maintaining their full control over the learning process. This approach aims to investigate the impact of SAM in fostering active participation and deeper engagement during independent learning. We evaluated the effectiveness of SAM in two studies. In the pilot study, we tested SAM among university students with similar educational backgrounds. Our goal was to evaluate the tool within a specific demographic, namely young university students, to gather preliminary insights on its design and usefulness. We then incorporated this feedback to refine our tool before expanding the study to a broader audience. In the main study, we introduced SAM to a diverse audience recruited through an online platform, i.e., Prolific, to assess its effectiveness across a more diverse participant pool with varied learning backgrounds. \autoref{fig:p_m} represents the structural elements and the differences between the pilot and main studies. We adjusted the study design for the main study to address potential knowledge gaps among participants, as detailed below.

For the purpose of our user studies, we adapted SAM to simplify the user experience. The selected YouTube video and corresponding slides were uploaded into the platform. Participants were only required to click a ``Watch Lecture'' button to initiate the session, streamlining the process and focusing on the interaction with SAM.

\subsection{Study Design}
\begin{figure*}[t]
	\centering
	\subfloat[Exemplary view of the video watcher with disabled chat interface.]{\includegraphics[height=6.5cm, keepaspectratio]{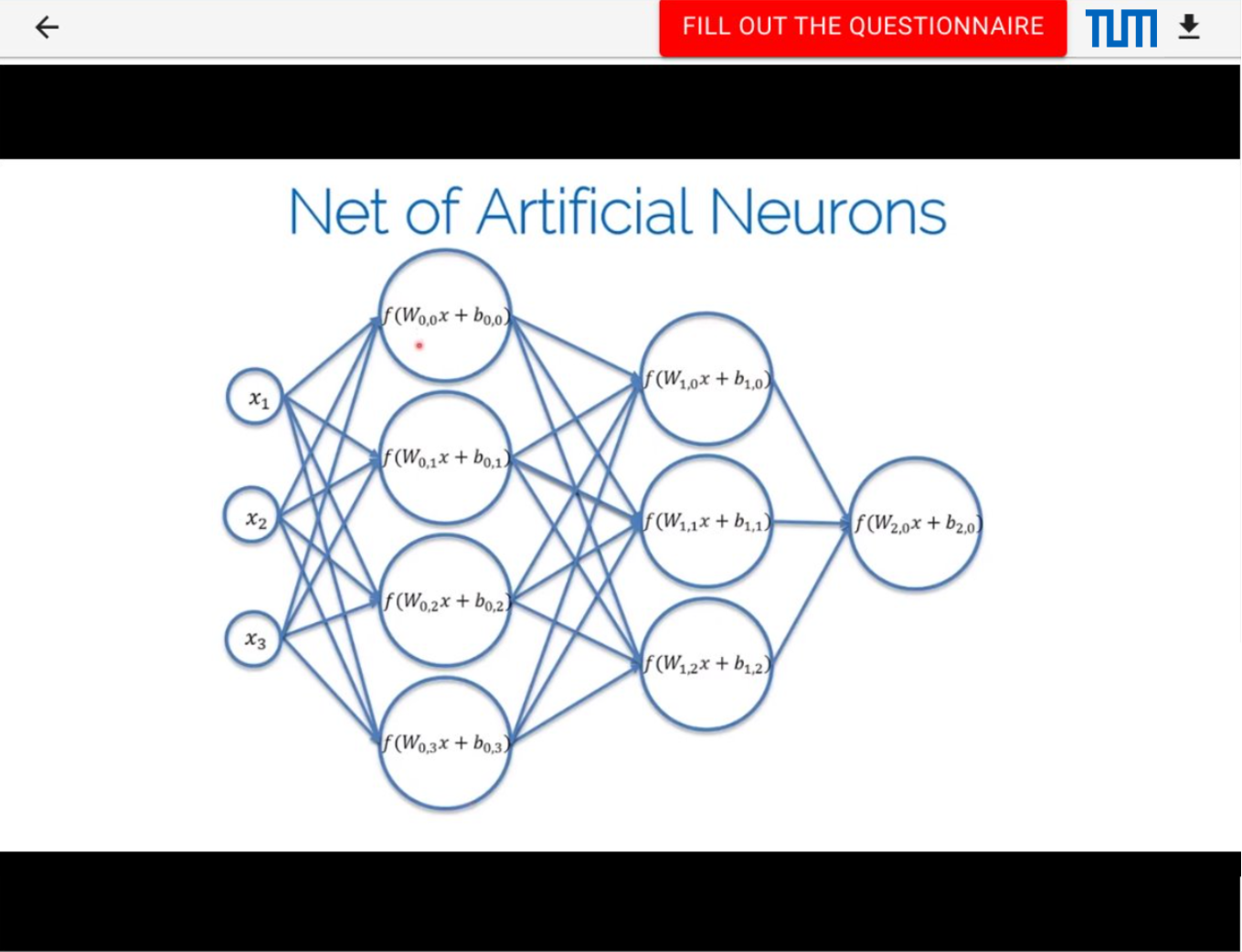}
    \label{fig:control_view}}
    \hfill % This will add flexible space between the images
	\subfloat[Exemplary view of SAM. At the bottom, users can start a conversation with the integrated AI mentor with the possibility of uploading an image.]{\includegraphics[height=6.5cm, keepaspectratio]{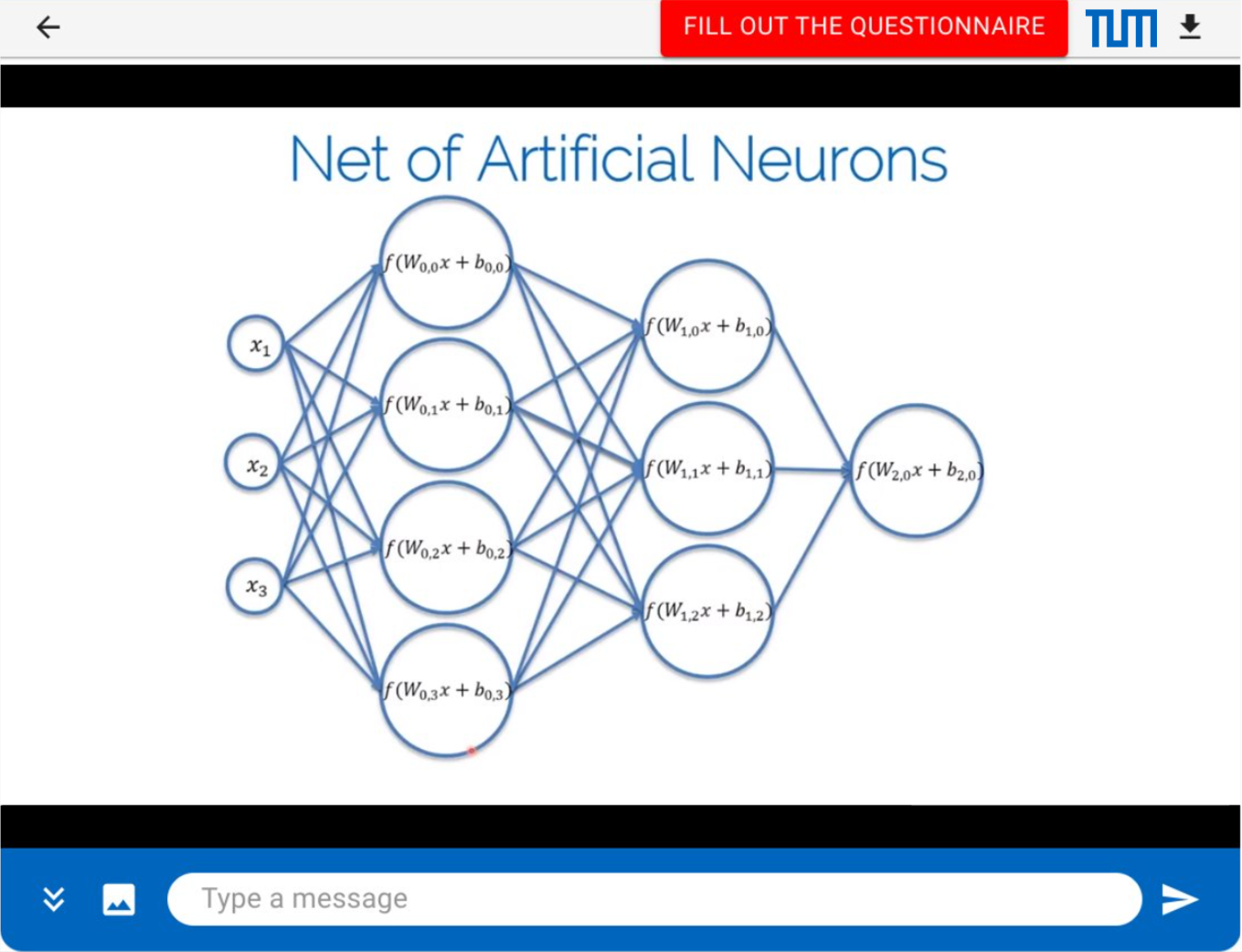}
	\label{fig:test_view}}
	\caption{Exemplary representations from the study.}
    \label{fig:study_views}
\end{figure*}

During both studies, we divided our participants into two groups: a test group and a control group. The test group had full access to SAM, meaning they could watch the video and interact with the AI mentor to ask questions anytime during the lecture (Figure~\autoref{fig:test_view}). In contrast, the control group also used SAM's video-watching interface, but the chat functionality was disabled (Figure~\autoref{fig:control_view}). This setup (\autoref{fig:study_views}) enabled us to compare the learning outcomes between participants who had access to the AI mentor and those who did not, providing insights into the impact of interactive AI support on the learning process. In the pilot study, participants were randomly assigned to groups using a backend function. For the main study, however, due to platform limitations (Prolific and Qualtrics), randomization was achieved by sequentially advertising separate study links, first the test group and then the control group. To ensure that participants could only enroll once, we restricted participation through Prolific. The advertisement included minimal information, primarily detailing exclusion criteria and specifically requesting that individuals with advanced knowledge of machine learning refrain from participating. After accepting the invitation, participants received additional information about the study.

After the pilot study, we made a few modifications to our study design. In the pilot study, all participants were enrolled in the same course and had attended the same previous 11 lectures, ensuring a similar background knowledge level across the group. However, the main study was advertised to a broader pool on Prolific, where participants varied in their prior knowledge. To account for this, we introduced pre-knowledge assessments to establish a baseline for each participant. We also included a brief theoretical overview, specifically an explanation of linear regression (refer to \autoref{chap:LR} for the full text), prior to the lecture. Additionally, based on insights from the pilot study, we standardized the test questions by ensuring each question had the same number of answer choices and an equal distribution of correct answers to more effectively measure knowledge gain. In the main study, we also introduced an attention check after 20 minutes of video playback to assess participants' attentiveness. A window popped up asking, ``Would you like to continue to watch the video?'' with a button labeled ``Continue.'' The time taken to click this button was recorded as an indicator of attentiveness.

When designing our user study, we drew inspiration from a study~\cite{alneyadi2023chatgpt} conducted in an elementary school where eleventh graders were divided into two groups: an experimental group with access to ChatGPT and a control group without it, using pre- and post-tests to measure learning outcomes; however, while their study was conducted in a traditional classroom setting over twelve weeks, our research focuses on the context of online learning.

After participants completed the lecture, a red button labeled ``Fill out the questionnaire'' appeared at the top right corner of the screen, directing them to the post-test questionnaire. This process differed between the test and control groups. While the control group only completed the post-test and demographics questionnaire, the test group had an additional questionnaire to provide feedback regarding their experience with SAM, offering insights into our tool's effectiveness and usability.

\subsection{Procedure}
This section outlines the three phases of our user studies, which were consistent across both the pilot and main studies. Each study followed the same sequence of phases to ensure comparability and reliability in the results. The differences between the pilot and main studies are highlighted in each phase and also summarized in \autoref{fig:p_m}. The questionnaires used in the studies were designed and administered through the Qualtrics platform.

\subsubsection{Phase 1: Pre-tests}
In the first phase of the pilot study, participants were asked to complete a pre-test (Appendix~\autoref{chap:pre_pilot}). The pre-test contained six questions and participants were informed that one or more answers could be correct, which helped measure their initial understanding. The pre-test questions were categorized into three key areas: graph neural networks, generative adversarial networks, and reinforcement learning, with two questions per category.

In the first phase of the main study, participants were asked to complete both a pre-knowledge test (Appendix~\autoref{chap:prek}) and a pre-test (Appendix~\autoref{chap:pre}). The pre-knowledge test, which consisted of six questions with only one correct answer each, was designed to assess and categorize participants based on their existing knowledge of neural networks (NNs). The pre-knowledge test was introduced to better understand participants' prior knowledge of NNs. This was not required in the pilot study, as it was conducted within a university course where participants had similar levels of familiarity with the topic. The pre-test also contained six questions with two correct answers each. The instructions remained consistent with the pilot study, indicating that one or more answers could be correct. Similarly, the pre-test questions were categorized into three key areas: the basics of NNs, the structure and training of NNs, and activation functions, with two questions per category. Following these assessments, participants were provided with a brief text on linear regression in the main study to ensure they had the necessary background knowledge before moving on to the video-watching phase of the study (Appendix~\autoref{chap:LR}).

\subsubsection{Phase 2: Lecture watching}
In the second phase of our studies, participants engaged in a lecture-watching activity. We piloted the platform among participants of a university course named as ``Introduction to Deep Learning'' at the Technical University of Munich, on the topic of deep learning. This pilot study focused on the \href{https://www.youtube.com/watch?v=DlHU4Ak-zjA}{12th lecture} of the course with a specific focus on more advanced topics, allowing us to gather preliminary feedback on SAM's usability and measure the participants' knowledge gain. The video had a duration of 1 hour and 56 minutes. As it encompassed the entire lecture content delivered by the instructor, it was not possible to reduce its length. We chose this lecture because NNs and machine learning are increasingly popular topics that resonate with a broad audience, making them ideal for engaging participants. The course at the university had hundreds of students enrolled, providing a substantial pool of potential participants. Participation in this pilot was entirely voluntary.

In the main study, participants watched the lecture titled \href{https://www.youtube.com/watch?v=Bdlj9uTtGDk}{``Introduction to Neural Networks''}. For the main study, we edited the \href{https://www.youtube.com/watch?v=eBeNzS1tAO4}{original video} down to 28 minutes, focusing exclusively on the introduction section. This adjustment was made to accommodate the diverse participant pool by ensuring sustained attention and engagement. Unlike the full-length lecture used in the pilot study, which spanned the entire university course session, the shorter video was better suited to the objectives of this study and the varied backgrounds of the participants. This approach also allowed us to streamline the content, making it more accessible and relevant for the study's purposes.
We chose this specific lecture because it introduces the fundamentals of NNs, a topic of growing interest in the field of AI. To tailor the content to our study's needs, we edited the original video to focus exclusively on NNs, ensuring the material was challenging enough to test SAM's AI-driven mentoring capabilities while still being conceivable by beginners. 

\subsubsection{Phase 3: Post-tests}
In the third phase of our studies, participants were asked to complete a post-test (Appendix~\autoref{chap:post_pilot}, \autoref{chap:post}) and a feedback questionnaire on SAM’s usability (only for the test group), and share their demographic information. This phase followed the same structure in both studies. The post-test comprised six questions, mirroring the structure of the pre-test, with two questions per topic across the same three categories. Both the control and test groups were required to complete this test in the studies. Participants were informed that one or more answers could be correct, similar to the pre-test questionnaire. This assessment was designed to gauge any improvements in the participants' understanding of the material after watching the lecture, with a specific focus on whether the AI-driven mentoring of SAM facilitated enhanced learning outcomes.

Key aspects of the feedback questionnaire included the response time of SAM, the quality and reliability of its answers, and whether their comprehension of the three NN topics improved after interacting with SAM. Afterward, we collected demographics from both groups, covering age, gender, and employment status.

\subsection{Participant Recruitment}
In our main user study evaluating SAM, we recruited participants from the Prolific platform, splitting them into two groups—one utilizing an AI mentor during video watching (i.e., test group) and the other simply watching the video (i.e., control group). We chose Prolific due to its reputation for providing high-quality data~\cite{Peer_etal_2022}. Our sample was gender-balanced, including participants who were 18 or older, located in Germany, and fluent in English. After recruitment, participants were directed to Qualtrics for the study, and upon completion, they returned to Prolific for compensation, which was set at a rate of \texteuro12.41/hour. Given the time difference between interacting with the AI mentor and watching the video uninterrupted, the main study duration varied between 42 and 45 minutes. All data was collected anonymously, with digital informed consent obtained from each participant before the study began. Participants could withdraw at any time without further consequences to them. For the main study on Prolific, we recruited $n_{control} = 51$ for the control group and $n_{test} = 59$ for the test group. For the pilot study in the university course, we recruited $n_{control} = 15$ for the control group and $n_{test} = 15$ for the test group.  

We excluded participants who completed the study in an unusually short amount of time. Given the time differences between the two groups and after analyzing outliers based on the total study duration, we set minimum time thresholds of 20 minutes for the control group and 35 minutes for the test group. These thresholds were consistent with testing in smaller groups, which identified the optimal time range for each phase. Participants who spent less time than this likely did not fully engage with the questions or the lecture, making their responses less reliable. We excluded participants who failed the attention check, which involved responding to a message that appeared during the video. They were excluded from the study if they did not click on this message within eight minutes. Additionally, participants who provided low-quality answers were excluded, as they directly entered the post-test questions into SAM, resulting in responses that did not accurately reflect their comprehension. After filtering out participants with age discrepancies greater than three years or mismatches in gender (e.g., listed as a man in one place and woman in another) between Prolific (if applicable) and Qualtrics, we ended up with the final sample sizes that are presented in \autoref{tab:sample_sizes}. We also provide more detailed demographics in the Appendix~\autoref{chap:dem}. Ethical considerations were strictly adhered to, with all user data collected anonymously, and an electronic consent obtained for storing, using, and potentially sharing the conversational data for research purposes.

\begin{table}[t]
\centering
\caption{Sample sizes and demographic information for pilot and main studies.}
\label{tab:sample_sizes}
\begin{tabular}{ccccccc}
\toprule
\textbf{Study}       & \textbf{Group} & \textbf{n} & \textbf{Age}  & \textbf{Men} & \textbf{Women} & \textbf{Diverse} \\ 
\midrule
\multirow{2}{*}{Pilot} & control & 11 & $23.9 \pm 2.3$ & 9 & 2 & 0 \\
                            & test & 14 & $23.4 \pm 1.9$ & 11 & 2 & 1 \\ 
\midrule
\multirow{2}{*}{Main}  & control & 38 & $29.9 \pm 5.8$ & 20 & 18 & 0 \\
                            & test   & 42 & $28.4 \pm 5.3$ & 21 & 21 & 0 \\
\bottomrule
\end{tabular}
\end{table}

\subsection{Analysis} \label{chap:analysis}
We evaluated the collected results through a comprehensive analysis of multiple aspects, including knowledge gain, user satisfaction, and answer accuracy. By examining these key metrics, we aim to assess the overall effectiveness of SAM in enhancing the learning experience. We assessed participants' overall performance by assigning points to each question based on the number of correct answers it contained. Participants earned points for correctly selecting valid answers and for refraining from selecting incorrect ones. We initially analyzed the total points achieved by participants in both the pre- and post-test phases. Additionally, we examined the number of incorrect selections made, providing insight into participants' accuracy and comprehension before and after using SAM. After the pilot study, a pre-knowledge test was included because self-reported assessments of prior knowledge, rated on a 5-point scale from ``no knowledge'' to ``expert knowledge,'' can be subjective and inconsistent. To obtain a more accurate measure of participants' understanding of NNs, we included six objective pre-knowledge questions in the main study. These questions provided a standardized basis for grouping participants according to their actual level of expertise, ensuring a more reliable analysis of the results. Participants who scored 1-2 points on the pre-knowledge test were categorized as having no prior knowledge, while those who scored 3-4 points were considered to have basic knowledge, and those who scored 5-6 points were classified as having intermediate knowledge. 

\begin{table*}[t] 
    \centering
    \caption{Analytical data and performance analysis from the pilot and main studies.}
    \begin{tabular}{p{34mm} p{12mm} p{12mm} p{12mm} p{12mm} p{12mm} p{12mm} p{12mm} p{12mm}}
    \toprule
        & \multicolumn{4}{c}{\textbf{Pilot Study}} & \multicolumn{4}{c}{\textbf{Main Study}}  \\
        & \multicolumn{2}{c}{\textbf{Control}} & \multicolumn{2}{c}{\textbf{Test}} & \multicolumn{2}{c}{\textbf{Control}} & \multicolumn{2}{c}{\textbf{Test}}  \\
        & \textbf{Pre} & \textbf{Post} & \textbf{Pre} & \textbf{Post} & \textbf{Pre} & \textbf{Post} & \textbf{Pre} & \textbf{Post}\\
        \midrule
        avg. pre-knowledge points & - & - & - & - & $3.6 \pm 1.2$ & - & $4.0 \pm 1.2$ & - \\
        avg. collected points & $17.5 \pm 3.1$ & $23.5 \pm 1.6$ & $18.4 \pm 2.6$ & $23.9 \pm 2.7$ & $13.1 \pm 2.5$ & $14.7 \pm 2.9$ & $14.8 \pm 2.9$ & $14.9 \pm 2.5$ \\
        avg. mistakes & $2.6 \pm 1.3$ & $1.3 \pm 0.9$ & $2.6 \pm 1.6$ & $1.4 \pm 1.8$ & $4.0 \pm 2.0$ & $3.2 \pm 1.5$ & $3.2 \pm 1.5$ & $2.5 \pm 1.3$ \\
    \bottomrule
    \end{tabular}
    \label{tab:points}
\end{table*}

In the pilot study, the maximum available points varied based on the number of correct answers per question across six pre-test and post-test questions, whereas in the main study, each of the six questions in both the pre-test and post-test featured four answer options with two correct answers each, standardizing the point distribution. In the pilot study, the pre- and post-test questions were validated by an instructor of the course. The difficulty level of the pre- and post-test questions in the main study was evaluated by 10 unrelated machine learning experts, who rated the questions on a scale of 1 to 3, with 1 standing for easy and 3 standing for  difficult. The average difficulty level for both the pre- and post-tests was 1.65, indicating that the questions were optimized for participants with little to no prior knowledge of the field, while still aiming to challenge their understanding. In both the pilot and main studies, the pre- and post-test questions varied, but consistently addressed the same topics covered in the lecture.

To account for the small sample size and variance differences, we used the Baumgartner-Weiss-Schindler (BWS) test ~\cite{baumgartner1998nonparametric} to compare knowledge gains between the test and the control groups. This statistical analysis allowed us to assess whether the differences in knowledge gain were significant. The corresponding p-values are reported in the results section (\autoref{chap:results}), providing insight into the effectiveness of SAM in enhancing learning outcomes. In addition to assessing SAM's impact on knowledge gain, we are exploring which demographic groups—such as age and employment status—benefit most from its use. This investigation addresses a gap identified in a prior meta-analysis~\cite{wu2024ai}, which emphasized the importance of determining which learner groups are best suited for AI chatbot-supported learning. Furthermore, we analyzed the frequently asked questions to understand user focus areas and identify key learning challenges, alongside a detailed examination of all hallucinations generated by SAM. This included evaluating every hallucination individually as well as the accuracy of the responses, their potential causes, and how users reacted to these inaccuracies. Additionally, we collected user feedback on our tool’s usability and satisfaction using a 5-point Likert scale.

\section{Results} \label{chap:results}
In this section, we present the findings from our study, focusing on the evaluation of SAM's effectiveness as an educational tool. By comparing participants' performance in the test and control groups and in demographic subgroups, we aim to offer insights into how triggered interactions enhance learning outcomes. Additionally, we explore how these insights can inform the design of future educational applications for more effective learning experiences.

On average, participants in the main study interacted with SAM (lecture watching and/or chatting) for $31.71 \pm 10.91 $ minutes and $31.78 \pm 17.64$ minutes in the control and test groups, respectively, reflecting consistent usage patterns across groups. While the average duration in the test and control groups are similar, the difference in standard deviation indicates that SAM users tended to spend more time actively engaging with the system. In both groups, participants frequently used playback controls such as speed adjustments and rewinds, which accounts for variations in interaction duration. The interaction duration in the pilot study was more diverse due to students utilizing SAM to prepare for the exam and note-taking while stopping the video.

\subsection{Quantitative Findings}
To provide a clear understanding of the impact SAM had on participants' learning, we first present the detailed pre- and post-test scores. The exact mean values achieved by individuals, as detailed in ~\autoref{tab:points}, show the pre- and post-test results for both the test and control groups in the pilot and main studies. The results from the main study indicate that participants in the test group made fewer incorrect selections after watching the video with SAM. We further analyzed these findings by calculating knowledge gains to draw meaningful conclusions, ensuring a comprehensive evaluation of SAM's impact on learning outcomes.

\begin{figure*}[t]
	\centering
 \subfloat[Achieved knowledge gain in the pilot and main studies.]{\includegraphics[height=6.5cm, keepaspectratio]{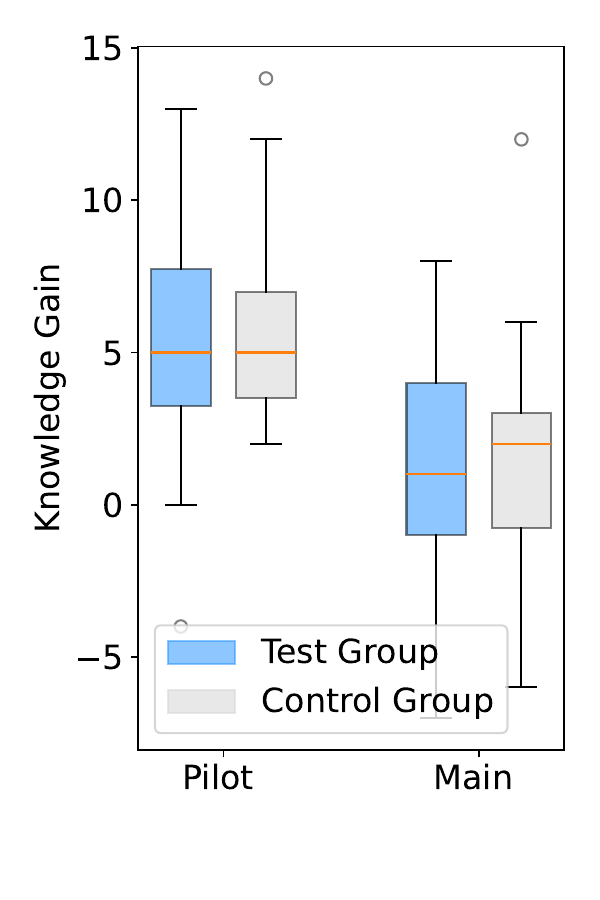}
    \label{fig:overall_kg}}
    \hfill
	\subfloat[Achieved knowledge gain in different age groups.]{\includegraphics[height=6.5cm, keepaspectratio]{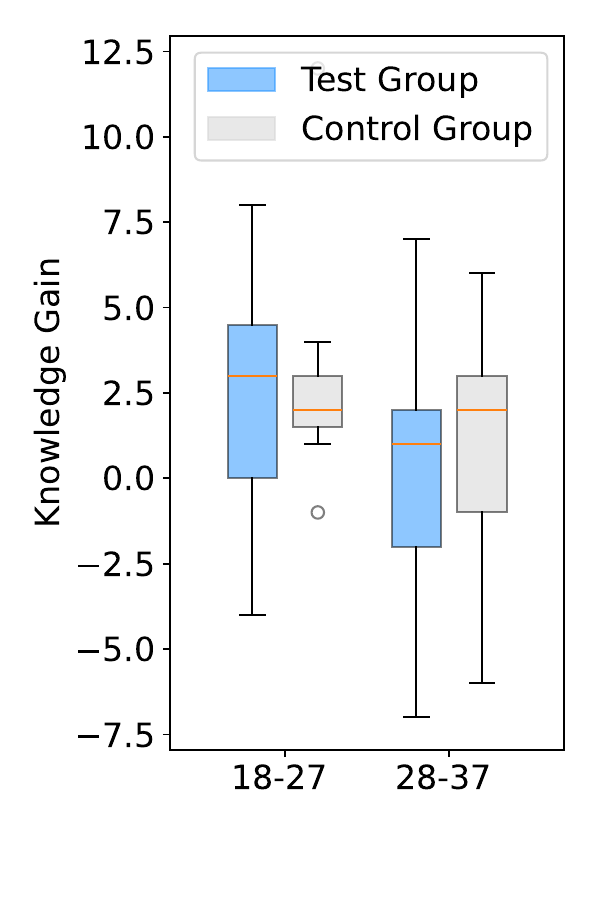}
    \label{fig:age}}
    \hfill % This will add flexible space between the images
	\subfloat[Achieved knowledge gain in different employment groups.]{\includegraphics[height=6.5cm, keepaspectratio]{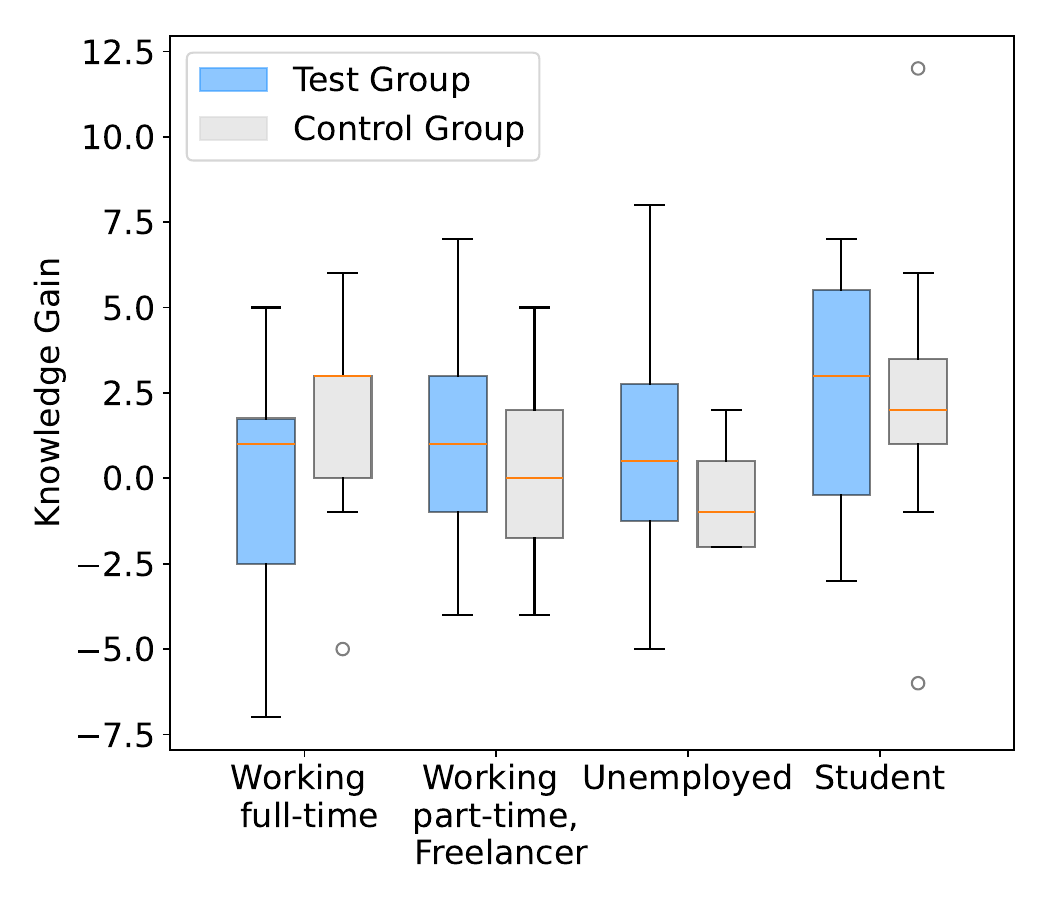}
	\label{fig:emp}}
	\caption{Box plots showing knowledge gains in test and control groups and across demographic subgroups.}
    \label{fig:kg_age_emp}
\end{figure*}

The statistical analysis of knowledge gains in both the pilot and main studies showed no significant differences between the test and control groups, as indicated by the high variance in the data (see Figure~\autoref{fig:overall_kg}). A comparison of the data distribution between the main and pilot studies revealed a significant difference with $p < .001$, suggesting that university students actively enrolled in a lecture employ different learning strategies compared to the diverse participant pool recruited via Prolific. To better understand the impact of SAM's features, we conducted a deeper analysis by examining the data across various subgroups, such as age and employment status. This involved comparing the performance of younger participants against older ones, as well as analyzing differences between students and full-time workers. Additionally, we looked at the test group in more detail, focusing on the number and nature of questions asked, which helped us identify how different participants engaged with SAM's context-aware capabilities.

To assess knowledge gains across different demographic groups, we focused on the main study due to the limited demographic diversity in the pilot study, where most participants were university students under 30 years old. The analysis of age groups revealed a significant difference ($p = .049$) in data distribution for participants under 28 years old between the test group (19 participants) and the control group (11 participants). Younger participants in the test group exhibited greater knowledge gains compared to their counterparts in the control group. Conversely, for participants aged 28–37 (21 in the test group, 23 in the control group), the trend was reversed, with the control group achieving higher knowledge gains ($p = .042$) (Figure~\autoref{fig:age}). Given the limited number of participants older than 37 (only two and four participants in the test and control groups, respectively), we focused only on the first two age groups to ensure representative data across all four subgroups. This trend suggests that our tool, SAM, which was primarily designed with students in mind, is particularly effective for younger learners, many of whom fall within the student demographic. This could indicate that younger participants, who are likely more accustomed to interactive learning technologies, benefit more from the AI-driven mentoring provided by SAM.

In our analysis of the main study, we examined knowledge gains in relation to employment status. We specifically considered groups with at least four participants to ensure meaningful comparisons. In Figure~\autoref{fig:emp}, we also represent part-time workers and freelancers, and unemployed, but based on their low representation, four and nine for part-time workers and freelancers, and four and four for unemployed participants in the control and test groups, respectively, we do not draw conclusions for these employment groups. The test group showed higher knowledge gain for students with 15 participants (mean of $1.8 \pm 3.6$) compared to the control group with 11 participants (mean of $1.4 \pm 3.9$) (Figure~\autoref{fig:emp}), although this difference is not statistically significant ($p > .05$), the observed trend and mean values provide support for our findings. However, our tool's effectiveness was less pronounced among those working full-time, whereas the control group performed better with 19 participants (mean of $1.9 \pm 2.7$), compared to the test group with 14 participants (mean of $0.1 \pm 3.3$) with a significance of $p = .022$. This finding suggests that students who have flexible work schedules can more easily engage with and benefit from SAM’s features, likely due to having more time to interact with the AI mentor, which is further supported by the observation that students in the test group spent, on average, more time interacting with SAM ($35.83 \pm 13.82$ minutes) compared to full-time workers ($30.58 \pm 24.65$ minutes). Our results are similar to the results from a previous study, which showed higher flexibility in learning enables learners to persist through challenges and adopt a wider array of cognitive and metacognitive strategies~\cite{bergamin2012relationship}. These results underscore SAM's potential as a valuable learning aid, particularly for students and other groups with adaptable learning environments.

\begin{figure*}[t]
	\centering
    \subfloat[The number of asked questions based on prior knowledge.]{\includegraphics[height=6.5cm, keepaspectratio]{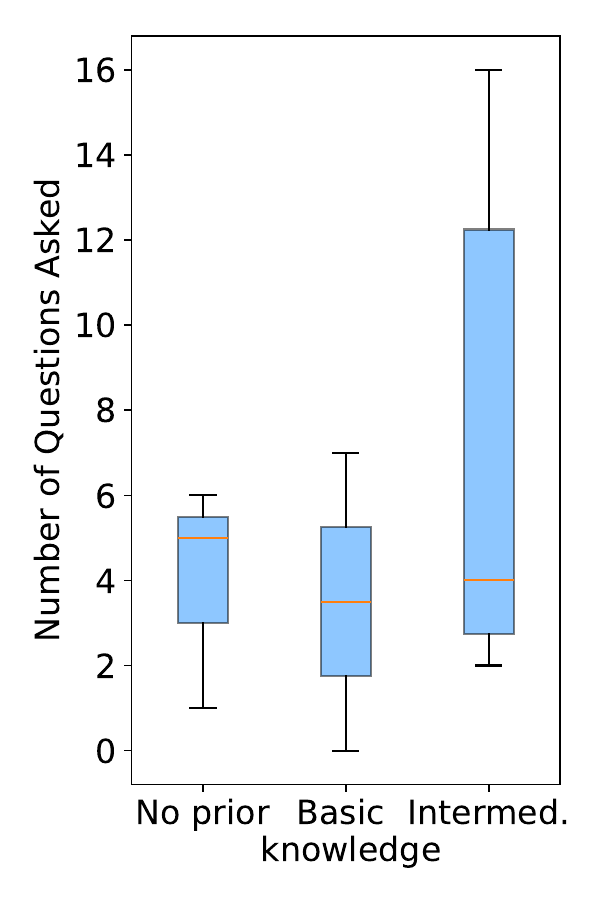}
    \label{fig:nn_q}}
     \hfil
	\subfloat[Distribution of participant questions over time during lecture watching, categorized by topic.]{\includegraphics[height=6.5cm, keepaspectratio]{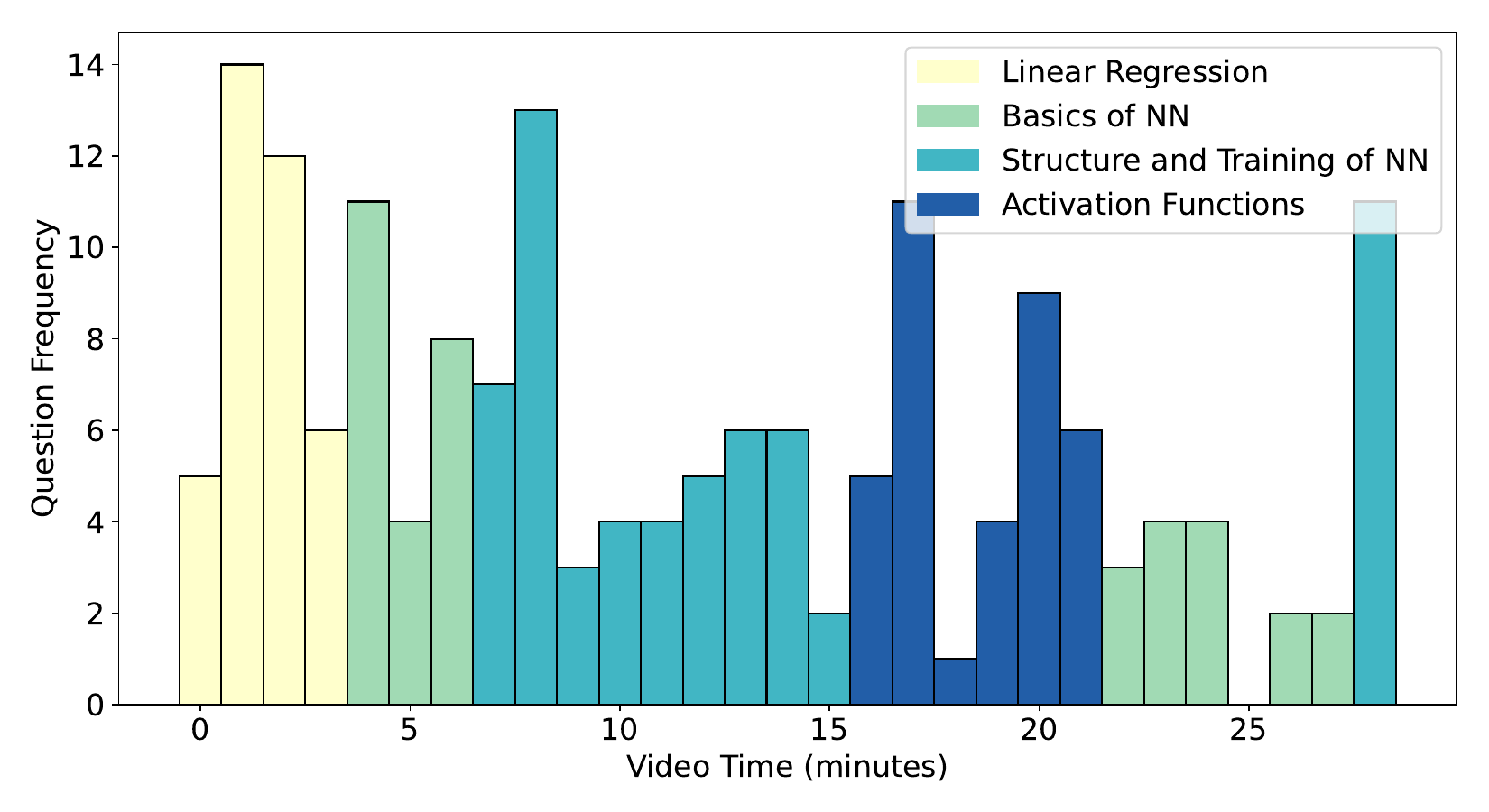}
	\label{fig:questions_over_time}}
	\caption{Analysis of the asked questions.}
    \label{fig:kg_3}
\end{figure*}

By further analyzing the data of the test group, Figure~\autoref{fig:nn_q} shows that participants with intermediate knowledge asked more questions (mean of $5.9 \pm 5.3$), than participants with basic knowledge (mean of $3.4 \pm 2.3$), although the distributions do not differ significantly ($p > .05$), a clear trend supports our findings. Based on the limited number of participants with no prior knowledge (six participants), we only made assumptions regarding participants with basic knowledge (20 participants) and intermediate knowledge (16 participants). The tendency for participants with prior knowledge to ask more questions further supports the idea that SAM is particularly useful for those looking to deepen their understanding, as these users were more likely to engage with the AI mentor to clarify and build upon the concepts presented in the lecture.

Additionally, we analyzed when these questions occurred during lecture watching. In Figure~\autoref{fig:questions_over_time}, we illustrate how often participants asked questions at each minute of the video, using different colors to represent the various topics. Notably, this figure presents the data user-wise, indicating the number of participants who asked at least one question at a given time point. A clear pattern emerged, with the highest number of questions being asked 2-3 minutes after a new topic was introduced. This suggests that participants took some time to process the material before seeking clarification. This observation highlights the importance of providing timely support during critical moments of learning, as participants are more likely to engage actively with the material once they have had time to familiarize themselves with a new concept. Also, a significant peak is observed at the last minute of the video, where participants asked for summaries or clarifications to reinforce their understanding before taking the post-test.

We also analyzed the frequently asked questions to better understand the areas where participants sought clarification or additional information while interacting with SAM. The three most common topics identified were weights in NNs, general information about NNs, and activation functions. To offer a thorough overview, we highlight the eight most frequently asked question categories, accompanied by two representative examples from each, detailed in the \autoref{chap:faq}. While most participants focused their questions on key topics and core concepts, a notable number inquired about basic elements, such as the  \href{https://www.cs.toronto.edu/%7Ekriz/cifar.html}{CIFAR-10} database, indicating limited prior knowledge in machine learning.

\subsection{Qualitative Feedback}

Overall, users in the pilot and main studies reported positive experiences with SAM, rating it as helpful and engaging on a 5-point Likert scale (see~\autoref{tab:feedback}), while a 3-point Likert scale was specifically used to assess whether participants avoided asking questions due to long response times. This is reflected in the generally high satisfaction scores $(3.4 \pm 0.7$ in the pilot and $4.0 \pm 0.5$ in the main study) and the willingness of participants to use SAM in the future ($3.6 \pm 0.7$ in the pilot and $4.3 \pm 0.8$ in the main study). In general, participants rated high response quality ($3.6 \pm 0.9$ in the pilot and $4.0 \pm 0.6$ in the main study) and trust in the accuracy of answers ($3.8 \pm 0.7$ in the pilot and $4.2 \pm 0.7$ in the main study), demonstrating SAM's effectiveness in providing real-time assistance. In both studies, participants found personalized learning support and feedback to be the most valuable feature of SAM. In the pilot study, this was followed by the ability to request summaries, receive formula explanations, and upload images to inquire about figures and formulas. In the main study, the order of preference shifted slightly, with formula explanations being the second most valued, followed by the ability to upload images for clarification on figures and formulas, and the option to request summaries.

Lastly, we evaluated the free-text feedback collected from participants, divided into two fields: feature suggestions and general messages to the developers. Suggestions included simplifying responses, improving the interface, enabling customizable interaction levels, adding voice recognition, providing visual aids, and integrating a video transcript with interactive explanations. Participants generally praised SAM as an innovative and helpful tool, especially for topics they had some prior knowledge of. They highlighted its ability to simplify concepts, generate formulas, and customize responses. While some found it less effective for entirely new topics, many appreciated its speed and functionality, with several expressing enthusiasm for its potential in education.

\begin{table}[t] 
    \centering
    \caption{User feedback from the pilot and main studies is presented as mean ratings with standard deviations on a 5-point Likert scale, except for ``Avoided asking due to delayed responses'', measured on a 3-point scale.}
    \begin{tabular}{c c c}
    \toprule
        \textbf{Feedback category} & \textbf{Pilot}  & \textbf{Main}  \\
        \midrule
        Satisfaction with Response time & $3.2 \pm 1.0$ & $4.2 \pm 0.7$ \\
        Avoided asking due to delayed responses & $1.3 \pm 0.5$ & $1.2 \pm 0.4$  \\
        Response quality & $3.6 \pm 0.9$ & $4.0 \pm 0.6$  \\
        Trust in accuracy of answer & $3.8 \pm 0.7$ & $4.2 \pm 0.7$  \\
        Enhanced learning experience & $3.5 \pm 0.5$ & $4.1 \pm 0.6$  \\
        Satisfaction & $3.4 \pm 0.7$ & $4.0 \pm 0.5$  \\
        Helpfulness & $3.5 \pm 0.9$ & $4.0 \pm 0.9$  \\
        Use in the future & $3.6 \pm 0.7$ & $4.3 \pm 0.8$  \\
    \bottomrule
    \end{tabular}
    
    \label{tab:feedback}
\end{table}

\subsection{AI tutor answer analysis}

\begin{table}[t] 
    \centering
    \caption{Participant responses to the question: ``Have you encountered any instances where SAM provided incorrect answers?''.}
    \begin{tabular}{c c c c c}
    \toprule
        \textbf{Study} & \textbf{Never} & \textbf{Occasionally}  & \textbf{Frequently} & \textbf{Unnoticed}  \\
        \midrule
        Pilot & 7 & 2 & 0 & 4 \\
        Main  & 23 & 2 & 2 & 15 \\
    \bottomrule
    \end{tabular}
    
    \label{tab:trust_sam}
\end{table}

In both the pilot and main studies, we asked participants to report whether they encountered any incorrect answers provided by SAM. Most participants indicated that they did not find or notice any inaccuracies. This feedback suggests that participants generally perceived SAM's responses as accurate and reliable. The detailed responses regarding the perceived accuracy of SAM's answers are summarized in \autoref{tab:trust_sam}, further supporting the tool's effectiveness in delivering correct and contextually relevant information.

Since the lecture material was mostly new to the participants, their responses regarding whether they encountered incorrect answers may not be fully reliable, necessitating a manual evaluation of SAM's provided answers to ensure accuracy. In the main study, we collected all user-generated questions and had them evaluated by two machine learning experts to assess the correctness of the responses provided by SAM. Two experts carefully reviewed each answer, considering not only its general correctness but also its relevance to the specific context of the lecture material. For instance, SAM occasionally provided accurate but contextually inappropriate answers, such as offering a general explanation of matrix indices when the question was specifically related to the indices within a neural network's weight matrix, a concept tied closely to the lecture's content on network structure. Because the majority of the provided answers were correct, resulting in unbalanced labels, Cohen's $\kappa$ cannot accurately reflect the inter-rater reliability. The experts achieved 96.0\% agreement, and after discussion, they found that the LLM provided a factually correct answer in 97.6\% of the cases. This high level of accuracy reflects SAM's effectiveness in responding to user queries within the context of the lecture material. A more detailed analysis, focusing on the factual correctness in each category, is provided in \autoref{tab:question}.

\begin{table*}[t] 
    \centering
    \caption{Analysis of the answers provided by SAM in the main study with respect to the three main categories.}
    \begin{tabular}{c c c c c}
    \toprule
         & \textbf{Basics of NN} & \textbf{Structure \& Training of NN}  & \textbf{Activation functions} & \textbf{Other}  \\
        \midrule
        num. of correct answers & 60 & 95 & 40 & 47 \\
        num. of incorrect answers & 0 & 4 & 2 & 0 \\
        Correctness rate & 100\% & 97.0\% & 96.0\% & 97.9\% \\
    \bottomrule
    \end{tabular}
    \label{tab:question}
\end{table*}

We further analyzed the hallucinations that the experts marked and how the users reacted to them. We analyzed each and every one of them individually. Most hallucinations did not contain outright false information but were often unrelated to the lecture material, which could still mislead users. In one case, the LLM responded with ``Please focus on the lecture material.'' instead of answering the question. On one occasion, the mistake was corrected in the following message. Overall, the majority of the responses did not contain factual errors. Interestingly, users never directly addressed or highlighted errors in their messages. For factual inaccuracies, it is likely that users did not realize the information was incorrect. We collected all hallucinations with explanations in the Appendix~\autoref{chap:hallucinations}.

\section{Discussion and Limitations}
SAM is an AI-driven educational tool designed to enhance learning by integrating real-time, context-aware question-answering with video content. A main user study involving 80 participants evaluated SAM's effectiveness, focusing on knowledge gain, user satisfaction, and answer accuracy. However, variability in participants' prior knowledge, engagement levels, or motivation could have influenced the results. In our main study, we used a video lecture from a university course as the main instructional material. This video, being the third in the course series, may not have been sufficient to bring all participants to a similar baseline understanding of the topic at hand. By providing a summary of the concept to the participants before watching the lecture helped address this issue. This gap is evident in the test group, where many participants sought further explanations (e.g., on linear regression) during the study. Although we instructed participants not to use external resources during the tests, we cannot be entirely certain that this was adhered to. We attempted to mitigate this by monitoring the time participants spent on the study.

Moreover, because the pilot study was integrated into the course, we lacked clear insights into the participants' prior knowledge. Since the question, ``Do you have previous knowledge in the following areas: Graph Neural Networks, Generative Models, or Reinforcement Learning?'' relies on self-reporting, we were unable to objectively measure how much they had absorbed from previous lectures or related courses in machine learning and deep learning offered by the same university. Feedback from the pilot study was valuable and informed adjustments for the main study, though some areas for improvement may have remained unaddressed. Another limitation identified was SAM's inconsistent performance in answering questions. In rare cases, SAM responded with ``I don't know,'' even when it had previously provided correct answers to differently phrased questions. There were also some instances where SAM failed to understand the specific context of a question, offering general but irrelevant answers to the current lecture content.

In designing SAM, we aligned its functionality with the ICAP framework~\cite{chi2014icap}, ensuring it operates at the interactive learning level, where learners actively engage through real-time interactions, such as they could discuss the given justifications, ask questions, and discuss similarities and differences if needed. In contrast, the control group was limited to the active learning level, allowing them only to manipulate the video without additional support. However, not all participants utilized the option to interact with the video, resulting in some engaging solely in passive learning. Future work could explore alternative support for the control group, such as a blog for asynchronous assistance. However, this approach was beyond the scope of this study due to practical constraints with Prolific, as varying interaction times would disrupt the real-time scenario we aimed to simulate. SAM’s primary goal is to provide immediate, real-time support, and offering instant responses through a blog would compromise the integrity of the results. Also, future work could explore how SAM's design principles inform broader educational theories, particularly in asynchronous or hybrid learning contexts.

Furthermore, while SAM was specifically designed with students in mind and demonstrated effectiveness for this subgroup, it did not perform as well for full-time employees. The structure of the tool emphasizes flexibility and interactive engagement, and it aligns well with the learning habits and needs of students who are accustomed to using digital platforms for their studies. However, full-time employees, who may have different learning preferences or time constraints, did not benefit from SAM as much. These findings suggest that while SAM is a powerful tool for enhancing student learning, its design might need to be adjusted to accommodate different user groups better, such as full-time professionals who require more tailored educational resources.

The evaluation in our study focused on a single topic, neural networks, which is highly structured and well-suited for AI-driven mentoring. The tool's applicability to disciplines requiring more abstract or discussion-oriented learning remains unexplored. Testing SAM in these diverse areas could provide insights into its versatility and effectiveness across different disciplines. In general, SAM's flexible design allows for adjustments to accommodate various subjects and lecture materials, making it adaptable to diverse educational contexts. Another limitation of SAM is the need for a thorough investigation into how lecture video quality impacts learning outcomes both with and without SAM's support. The present study relied on high-quality instructional videos with clear audio, well-organized content, and accompanying slides. However, the impact of varying video quality on SAM’s effectiveness has not yet been systematically examined. Factors such as audio clarity, visual fidelity, and the coherence of lecture materials may influence both user engagement and the AI’s ability to provide accurate, contextually relevant responses.

Additionally, although SAM is flexible in its use, it is currently optimized for videos available on YouTube. This is a limitation given that many educational institutions utilize other platforms, such as Panopto~\cite{getenet2022students}, for hosting and streaming lecture content. Future iterations of SAM should consider integrating with a broader range of video-watching tools to enhance its applicability and reach within academic environments. Expanding SAM’s compatibility with widely used educational platforms would not only increase its utility but also support its adoption across various learning contexts, ensuring that more students and educators can benefit from its capabilities.

While this study primarily measured short-term knowledge gains and user satisfaction, future research should assess SAM’s impact on long-term knowledge retention and higher-order cognitive skills, such as critical thinking and problem-solving. Metrics such as retention rates, frequency of subsequent interactions with SAM, and the transferability of learned concepts to novel scenarios would provide a more comprehensive evaluation of the tool’s educational value. Additionally, longitudinal studies that track learners’ progress over time could elucidate the sustained impact of SAM on academic achievement and self-directed learning behaviors. These investigations would not only validate SAM’s effectiveness but also contribute to the design of adaptive learning systems that promote lifelong learning.

Although SAM currently utilizes GPT-4o~\cite{gpt4o} as its underlying LLM, this model is easily exchangeable, allowing for flexibility in integrating other commercial or open-source LLMs. Our primary focus has been on SAM's long-term impact in fostering personalized, real-time learning support rather than being tied to a specific model. Investigating the most reasonable LLM for SAM's continued development is a future direction for this project. 

Overall, SAM introduces a novel approach to real-time support by delivering highly accurate answers combined with personalized learning experiences tailored to individual needs. This innovation showcases the potential of adaptive AI systems to enhance engagement and effectiveness in educational tools.

\section{Conclusion}
We introduced SAM, a context-aware AI chatbot that allows students to ask questions and explore unclear parts of the lecture. User studies with a total of 105 participants (80 in the main and 25 in the pilot study) revealed its effectiveness as an educational tool focusing on proactive teaching. SAM enhances learning outcomes by encouraging students to take full ownership of their learning experience and fosters a more interactive and engaged learning environment. The real-time, personalized assistance provided by SAM allows learners to delve deeper into complex topics, reinforcing their understanding of the lecture content. The studies highlighted SAM's potential as an effective learning tool, and the detailed analysis of demographic groups revealed that SAM has a great potential, especially for young learners and individuals in flexible working environments, such as students, to increase knowledge gains. We also found that participants who had intermediate-level prior knowledge asked more questions during lecture watching. According to expert raters, SAM maintained a high answer correctness rate of 97.6\%, and overall, participants provided positive feedback regarding user satisfaction, response quality, and the tool's helpfulness, with many expressing a willingness to use SAM in the future. These findings underscore SAM's potential to transform education by promoting interactivity and learner independence. Future versions could expand its use beyond STEM subjects to areas that focus on discussion, interpretation, and critical thinking, while also incorporating features like summarizing argumentative texts to support learning in non-STEM disciplines.

\begin{acks}
We extend our gratitude to Prof. Dr. Matthias Nießner and his group for providing the lecture videos and allowing us to pilot our study in his course. We also thank students Rahmiye Büşra Büyükgebiz and Ozan Aydın for their invaluable contributions in developing the first prototype of SAM. Finally, we deeply appreciate Márton Szép for his insightful ideas, feedback, and assistance with writing.
\end{acks}

\bibliographystyle{ACM-Reference-Format}
\bibliography{main}

\appendix

\section{Appendix}

\subsection{Participant details} \label{chap:dem}

\autoref{tab:demographics} represents the detailed demographic distribution based on the questionnaires.
\begin{table*}[htbp] 
    \centering
    \caption{Detailed Demographics data.}
    \begin{tabular}{c c c c c c}
    \toprule
        & & \multicolumn{2}{c}{\textbf{Pilot Study}} & \multicolumn{2}{c}{\textbf{Main Study}}  \\
        & & \textbf{Control} & \textbf{Test} & \textbf{Control} & \textbf{Test}  \\
        \midrule
        Age &  & $23.9 \pm 2.3$ & $23.4 \pm 1.9$ & $29.4 \pm 5.8$ & $28.4 \pm 5.3$ \\
        \midrule
        \multirow{3}{*}{Gender}& Man & 9 & 11 & 20 & 21 \\
        & Woman & 2 & 2 & 18 & 21 \\
        & Diverse & 0 & 1 & 0 & 0 \\
        \midrule
        & Working full-time & 0 & 0 & 19 & 14 \\
        & Working part-time & 2 & 1 & 1 & 7 \\
        Employment & Unemployed & 0 & 0 & 4 & 4 \\
        & Freelancer & 0 & 0 & 3 & 2 \\
        & Student & 9 & 13 & 11 & 15 \\
        \midrule
        \multirow{3}{*}{Prior ML}& No experience & 6 & 4 & 17 & 14 \\
        & User & 5 & 8 & 19 & 27 \\
        & Programmer & 0 & 2 & 2 & 1 \\
    \bottomrule
    \end{tabular}
    
    \label{tab:demographics}
\end{table*}

\subsection{Frequently asked questions} \label{chap:faq}

Here we present examples of frequently asked questions that highlight the most common topics of interest among participants. These examples provide insight into the areas where users sought additional clarification or deeper understanding while interacting with SAM.

\begin{itemize}
    \item Linear regression:
        \begin{itemize}
            \item ``What is linear regression?''
            \item ``Define "linear regression" in really easy terms''
        \end{itemize}
    \item Neural networks:
        \begin{itemize}
            \item ``what are neural networks in simple terms''
            \item ``Define a neural network''
        \end{itemize}
    \item Weights:
        \begin{itemize}
            \item ``how to calculate the weight?''
            \item ``what is a weight matrix''
        \end{itemize}
    \item Hidden layer:
        \begin{itemize}
            \item ``what is a hidden layer''
            \item ``what is the hidden layer?''
        \end{itemize}
    \item Activation function:
        \begin{itemize}
            \item ``what do activation fuctions do?''
            \item ``define activation function''
        \end{itemize}
    \item Tanh:
        \begin{itemize}
            \item ``what is a tanh function''
            \item ``what is tanh on the screen''
        \end{itemize}
    \item \href{https://www.cs.toronto.edu/%7Ekriz/cifar.html}{CIFAR-10} : (image database with 10 classes)
        \begin{itemize}
            \item ``what is cifar?''
            \item ``What is CIFAR-10?''
        \end{itemize}
    \item Summary:
        \begin{itemize}
            \item ``explain the whole video''
            \item ``please summarize the whole video shortly''
        \end{itemize}
\end{itemize}

\subsection{Hallucinations} \label{chap:hallucinations}
We conducted a detailed analysis of all hallucinations that occurred during user interactions with SAM. We documented the specific inaccuracies in the LLM's responses, providing commentary on why the given answer was incorrect. Additionally, we examined user reactions to these hallucinations, offering insights into how such errors impacted their experience and engagement with the system.

\begin{itemize}
    \item Question: is it possible to see a visual example of an output \textit{(Context: ``output'' refers to the output of a last layer in the neural network.)}
    
    Answer: Please focus on the lecture material.

    Incorrect: This was part of the lecture material.
    
    User reaction: ah nice, in this slide he's explaining what I've asked you before about the activation functions right? \textit{Comment: the participant did not ask further question on this topic, moved on with the lecture. No inaccuracies were stated by SAM.}

    \item Question: what are indices \textit{(Context: matrix indices for $W$ and $b$)}
    
    Answer: In the context of your lecture, indices are used to refer to specific elements within arrays, matrices, or tensors. Each index corresponds to a dimension of the array:

\begin{enumerate}
    \item \textbf{Single Index:} If you have a vector (a one-dimensional array), a single index will refer to a specific element in that vector. For example, if $ \mathbf{v}$ is a vector, $v_i$ would refer to the  \textit{i}-th element of $ \mathbf{v}$.
    \item \textbf{Multiple Indices:} When dealing with matrices (two-dimensional arrays) or higher-dimensional tensors, multiple indices are used. For example:
    \begin{itemize}
        \item For a matrix $ \mathbf{M}$, $M_{ij}$ refers to the element in the \textit{i}-th row and \textit{j}-th column.
        \item In neural networks, as mentioned in your transcript, a bias term $B$ might have two
    indices. The first index $B_{l}$ could indicate the layer in the neural network, and the
    second index $B_{l,n}$ could represent the specific neuron or node within that layer.
    \end{itemize}
\end{enumerate}

By using indices, you can refer to and manipulate specific elements within more complex
data structures, which is essential for performing operations like matrix multiplication and
applying transformations in neural networks.

    Incorrect: The answer itself is correct, however it is misleading related to the lecture. In NN, the matrix indices do not necessarily have two, sometimes only the layer is mentioned. 
    
    User reaction: define activation function \textit{(Comment: the participant did not ask further question on this topic, moved on with the lecture.)}

    \item Question: can you explain what max represents in this video \textit{(Context: The maximum function is commonly used in various activation functions.)}
    
    Answer: In simple terms, an activation function in a neural network is a rule that decides whether a particular neuron should be activated or not. Think of it as a decision-making step that determines if the input information is important enough to pass on to the next layer of the network. The activation function introduces non-linearity to the model, enabling the neural network to learn and model complex data patterns. Here's a very basic analogy: Imagine you are sorting emails, and your job is to decide whether each email should go to the ``Important'' folder or the ``Trash'' folder. The rules you use to make this decision (like checking if the email is from your boss or if the subject line contains certain keywords) are similar to what an activation function does. It evaluates the input and decides whether to pass the information along based on certain criteria. Would you like more details about the types of activation functions or their specific roles?

    Incorrect: This answer aligns with the lecture, as it correctly highlights that while activation functions are generally non-linear to enable learning complex patterns. However, the answer does not address the role of the maximum function in activation functions or provide specific examples.
    
    User reaction: The conversation stopped.

    \item Question: how do we estimate the value of the hidden layer?
    
    Answer: Estimating the value (or size) of the hidden layer in a neural network often involves a combination of theoretical understanding and empirical testing. Here are some common approaches:
\begin{enumerate}
    \item \textbf{Theoretical Guidelines:}
    \begin{itemize}
        \item \textbf{Input and Output Dimensions:} Consider the dimensions of your input and output. For example, if your input is high-dimensional, you might need a larger hidden layer to capture the complexity of the data.
        \item \textbf{Complexity of the Problem:} More complex problems might require larger or multiple hidden layers to learn the intricate patterns in the data.
    \end{itemize}
    \item \textbf{Empirical Methods:}
    \begin{itemize}
        \item \textbf{Cross-Validation:} Use techniques like cross-validation to test different sizes for the hidden layer and determine which size provides the best performance on validation data.
        \item \textbf{Grid Search:} Perform a grid search over a range of values for the hidden layer size to systematically evaluate performance.
    \end{itemize}
    \item \textbf{Common Practices:}
    \begin{itemize}
        \item A common starting point is to use hidden layer sizes that are between the input size and the output size. For instance, if your input layer has 16384 units and your output layer has 10 units, you might start with hidden layers having sizes anywhere from a few hundred to a few thousand units.
        \item Use powers of 2: Sizes such as 128, 256, 512, 1024, etc., are frequently used.
    \end{itemize}
    \item \textbf{Regularization and Dropout:}
    \begin{itemize}
        \item When using larger hidden layers, it is essential to incorporate regularization techniques like dropout, L2 regularization, etc., to prevent overfitting.
    \end{itemize}
\end{enumerate}

In the context of the lecture, a hidden layer was given an arbitrary dimension, such as 1000,
which is a reasonable choice for experimentation. The exact size would ideally be determined through empirical testing and cross-validation to find the optimal configuration for your specific task.

    Incorrect: It seems likely that the participant was referring to the calculation of the dimensions of the hidden layer rather than the process of choosing its size. While the provided answer was technically correct, it addressed how to determine the size of a hidden layer, which may have been slightly misleading in the context of the question. A more precise explanation could help clarify the distinction between calculating dimensions and selecting the size based on specific criteria.
    
    User reaction: how to find the neuron to weight connection \textit{Comment: The participant continued the conversation, slightly changing the topic.}

    \item Question: what value does the scaler add to this equation? \textit{(Context: The participant is referring to the bias term.)}
    
    Answer: The scalar in the context of the lecture adds a bias term to the equation. Biases are critical in neural networks because they allow the model to better fit and approximate complex functions.Let's break it down with an example: Consider the output of a neuron in a neural network layer, which can be calculated as: $y = W \cdot x + b $
\begin{itemize}
    \item $W$ is the weight matrix,
    \item $x$ is the input vector,
    \item $b$ is the bias term (scalar),
    \item $y$ is the output.
\end{itemize}

The bias term $b$ is added after the dot product of $W$ and $x$. This addition shifts the activation function to better fit the data. Without a bias, the neuron would always output zero when all inputs are zero. The bias allows the activation function to be shifted left or right, which enables the neural network to fit the data more accurately. In the context of the video transcript:

``important when you're talking about these biases B has two indices so 0 0 means there's two indices and this is a scala that is added on top after you did this, this dot product here so this is a scaler and you have two indices here first one tells you the layer second one tells you the norm right this is a scalar this one is a vector and this one is also a vector right and doing a whole layer together is matrix vector multiplication plus a vector of biases'' The scalar bias component ensures that each neuron can output a non-zero value even when the input is zero, thereby enhancing the model's ability to learn complex data patterns.

    Incorrect: The activation function is missing from the formula.
    
    User reaction: sam, can you prepare an image for me of this structure with proper labelling of all componenets? \textit{Comment: Asking for further explanation, which contains the activation function.}

    \item Question: where else could someone use the MaxOut function? \textit{(Context: This is an introduced activation function in the lecture.)}
    
    Answer: The MaxOut function can be used in various scenarios beyond traditional neural networks. Here are a few examples:
\begin{enumerate}
    \item \textbf{Complex Neural Network Architectures:} MaxOut can be used in convolutional neural
networks (CNNs) for image recognition tasks, recurrent neural networks (RNNs) for sequence prediction tasks, and other deep learning models to enhance model performance and representational capacity.
\item \textbf{Ensemble Methods:} MaxOut can also be useful in ensemble methods where multiple
models' outputs are combined. By selecting the maximum output among several model predictions, MaxOut can help in making more robust predictions.
\item \textbf{Generative Models:} In generative models like Generative Adversarial Networks (GANs), MaxOut can be employed in both generator and discriminator networks to improve
the quality of generated samples.
\item \textbf{Automated Machine Learning:} In AutoML systems, MaxOut can serve as one of the
activation function options that the system can explore to find the best model configuration
automatically.
\item \textbf{Regularization Techniques:} MaxOut can act as a form of implicit regularization,
providing robustness against overfitting by its inherent property of choosing the maximum
activation.
\end{enumerate}

These are just a few examples, and the use of MaxOut can extend to any domain where neural networks and machine learning models are applied.

    Incorrect: The answer only collects NN related applications, without mentioning a broader field. The provided information was correct.
    
    User reaction: The conversation stopped.
    
\end{itemize}

\subsection{Linear Regression Summary from the Pre-Test Questionnaire} \label{chap:LR}

The following text appeared for each participant before watching the video:

``Linear regression is a statistical method used to model the relationship between a dependent variable and one or more independent variables by fitting a linear equation to the observed data. The simplest form, simple linear regression, involves a single independent variable and models the relationship with the equation $y = m \cdot x + b$, where \textit{y} is the predicted value, \textit{x} is the independent variable, \textit{m} is the slope of the line (indicating the relationship strength and direction between \textit{x} and \textit{y}), and \textit{b} is the y-intercept (the value of \textit{y} when \textit{x} is zero). In multiple linear regression, the model includes multiple independent variables and extends to
\begin{equation}
    y = b_0 + b_1 \cdot x_1 + b_2 \cdot x_2 + ... + b_n \cdot x_n.
\end{equation} 

The coefficients $b_i$ are estimated using methods such as Least Squares, which minimizes the sum of the squared differences between the observed values and the values predicted by the model. The goal is to find the best-fitting line that explains the variation in the dependent variable based on the independent variables, enabling predictions of \textit{y} for given values of \textit{x}.

Linear regression connects to classification tasks by providing a foundation for understanding how inputs (such as images of cats and dogs) can be used to predict outputs (labels - "cat" or "dog"). While linear regression predicts continuous values, classification tasks involve predicting discrete class labels. Techniques similar to linear regression, such as logistic regression, extend this concept by applying a transformation to the linear equation to handle binary or multiclass classification. This approach models the probability that a given input belongs to a particular class, enabling the assignment of class labels based on the highest predicted probability.''

\subsection{Questionnaires - Pilot Study}
\subsubsection{Pre-Test} \label{chap:pre_pilot}
\textbf{Topic 1: Graph Neural Networks}

\begin{enumerate}
    \item  In a graph neural network, what do nodes and edges represent?
    \begin{itemize}[label=$\square$]
        \item Nodes represent features, and edges represent labels.
        \item Nodes represent computational units, and edges represent the flow of information or dependencies. 
        \item Nodes represent latent variables, and edges represent probabilistic dependencies.
        \item \textit{Nodes represent entities with attributes, and edges represent the connections or correlations among them.} \cmark
    \end{itemize}

    \item What are significant challenges when working with graph neural networks?
    \begin{itemize}[label=$\square$]
        \item Handling fixed-size input data.
        \item \textit{Ensuring invariance to node permutations.} \cmark
        \item Avoiding the use of activation functions.
        \item \textit{Capturing the complex relationships in graph-structured data.} \cmark
    \end{itemize}
\end{enumerate}

\textbf{Topic 2: Generative Adversarial Networks}

\begin{enumerate}[resume]
    \item Which of the following statements correctly describe the loss functions used in training Generative Adversarial Networks (GANs)?
    \begin{itemize}[label=$\square$]
        \item The generator's loss aims to minimize the Kullback-Leibler (KL) divergence between the generated data distribution and the real data distribution.
        \item \textit{The discriminator's loss aims to maximize the likelihood of correctly distinguishing between real and generated data.} \cmark
        \item \textit{The generator's loss typically includes a term that maximizes the probability of the discriminator being incorrect.} \cmark
        \item The discriminator's loss is computed using the mean squared error between the real and generated data labels.
        \item \textit{Both generator and discriminator use cross-entropy loss functions in their original formulation.} \cmark
    \end{itemize}

    \item Which of the following are key characteristics of generative models?
    \begin{itemize}[label=$\square$]
        \item They require labeled data for training.
        \item \textit{They generate new samples from the same distribution as the training data.} \cmark
        \item \textit{They involve optimization techniques.} \cmark
        \item They are only used for regression tasks.
    \end{itemize}
\end{enumerate}

\textbf{Topic 3: Reinforcement Learning}

\begin{enumerate}[resume]
    \item Which of the following best describes the difference between supervised learning, unsupervised learning, and reinforcement learning?
    \begin{itemize}[label=$\square$]
        \item \textit{Supervised learning uses labeled data, unsupervised learning uses unlabeled data, and reinforcement learning uses reward signals to learn.} \cmark
        \item Supervised learning uses reward signals, unsupervised learning uses labeled data, and reinforcement learning uses unlabeled data.
        \item Supervised learning uses unlabeled data, unsupervised learning uses labeled data, and reinforcement learning uses both labeled and unlabeled data.
        \item Supervised learning uses labeled data, unsupervised learning uses reward signals, and reinforcement learning uses unlabeled data.
    \end{itemize}

    \item In reinforcement learning, what is a policy?
    \begin{itemize}[label=$\square$]
        \item A function that maps actions to states.
        \item \textit{A function that maps states to actions.} \cmark
        \item A reward signal over time.
        \item A state transition model.
    \end{itemize}
\end{enumerate}

\subsubsection{Post-Test} \label{chap:post_pilot}
\textbf{Topic 1: Graph Neural Networks}

\begin{enumerate}
    \item  Which of the following are applications of graph neural networks?
    \begin{itemize}[label=$\square$]
        \item Autonomous vehicle navigation
        \item \textit{Protein structure prediction} \cmark
        \item Stock market prediction
        \item \textit{Social network analysis} \cmark
    \end{itemize}

    \item In graph neural networks, what does the process of message passing involve?
    \begin{itemize}[label=$\square$]
        \item Passing fixed-size vectors through convolutional layers.
        \item \textit{Aggregating information from neighboring nodes to update node representations.} \cmark
        \item Using recurrent units to maintain state information.
        \item Applying dropout for regularization.
    \end{itemize}
\end{enumerate}

\textbf{Topic 2: Generative Adversarial Networks}

\begin{enumerate}[resume]
    \item Which of the following are generative models?
    \begin{itemize}[label=$\square$]
        \item \textit{Variational Autoencoder} \cmark
        \item Support Vector Machine
        \item \textit{Generative Adversarial Network} \cmark
        \item \textit{Fully Visible Belief Nets} \cmark
        \item Convolutional Neural Network
        \item \textit{Boltzmann Machine} \cmark
    \end{itemize}

    \item What are the primary differences between Variational Autoencoders (VAEs) and Generative Adversarial Networks (GANs)?
    \begin{itemize}[label=$\square$]
        \item VAEs use a discriminator to distinguish real from fake data, while GANs do not.
        \item \textit{VAEs optimize a likelihood function, while GANs use a minimax game between generator and discriminator.} \cmark
        \item VAEs require labeled data for training, while GANs do not.
        \item \textit{VAEs explicitly model the data distribution, while GANs learn to generate data through adversarial training.} \cmark
    \end{itemize}
\end{enumerate}

\textbf{Topic 3: Reinforcement Learning}

\begin{enumerate}[resume]
    \item Which of the following concepts are fundamental to reinforcement learning?
    \begin{itemize}[label=$\square$]
        \item \textit{Reward} \cmark
        \item Supervised learning
        \item \textit{Policy} \cmark
        \item Loss function
        \item \textit{Environment} \cmark
        \item Gradient Descent
    \end{itemize}

    \item Which of the following best describes the Markovian assumption in reinforcement learning?
    \begin{itemize}[label=$\square$]
        \item \textit{The future state depends only on the current state and action, not on the sequence of events that preceded it.} \cmark
        \item The future state depends on the entire history of states and actions.
        \item The future reward depends on the initial state of the environment.
        \item The future state and reward are independent of the current action.
    \end{itemize}
\end{enumerate}

\subsection{Questionnaires - Main Study}
\subsubsection{Pre-Knowledge Test} \label{chap:prek}
\begin{enumerate}
    \item What does ‘overfitting’ mean in the context of machine learning?
    \begin{itemize}[label=\textcircled{}]
        \item When a model performs poorly on the training data
        \item When a model performs poorly on the testing data
        \item When a model performs equally well on both training and testing data
        \item \textit{When a model performs well on the training data but poorly on new, unseen data} \cmark
    \end{itemize}
    
    \item Which algorithm is commonly used for training neural networks?
    \begin{itemize}[label=\textcircled{}]
        \item \textit{Gradient Descent} \cmark
        \item K-means Clustering
        \item Principal Component Analysis
        \item Apriori Algorithm
    \end{itemize}
    
    \item What is training in machine learning?
    \begin{itemize}[label=\textcircled{}]
        \item Teaching people to use computers
        \item \textit{Improving a model with data} \cmark
        \item Writing a program to solve specific tasks
        \item Eliminating mistakes from a computer program
    \end{itemize}
    
    \item In neural networks, what is the purpose of an activation function?
    \begin{itemize}[label=\textcircled{}]
        \item To initialize the weights
        \item \textit{To transform the input signal into an output signal} \cmark
        \item To measure the error of the model
        \item To update the weights during backpropagation
    \end{itemize}
    
    \item What is a dataset?
    \begin{itemize}[label=\textcircled{}]
        \item A set of computer programs
        \item A type of database system
        \item \textit{A collection of data for training} \cmark
        \item Information available on Google
    \end{itemize}
    
    \item What is the primary goal of supervised learning?
    \begin{itemize}[label=\textcircled{}]
        \item To find patterns in unlabeled data
        \item To generate new data
        \item To perform dimensionality reduction
        \item \textit{To find patterns in labeled data} \cmark
    \end{itemize}
\end{enumerate}

\subsubsection{Pre-Test} \label{chap:pre}
\textbf{Topic 1: Basics of Neural Networks}

\begin{enumerate}
    \item What are the main goals of a machine learning model?
    \begin{itemize}[label=$\square$]
        \item \textit{Fit a function to the training data} \cmark
        \item To approximate the data distribution 
        \item Optimize for the highest accuracy on the training set
        \item \textit{Generalize from the training set} \cmark
    \end{itemize}

    \item What roles do non-linearities play in a neural network?
    \begin{itemize}[label=$\square$]
        \item They enhance computational efficiency
        \item \textit{They increase model complexity} \cmark
        \item They ensure a more straightforward optimization process
        \item \textit{They allow the network to approximate any continuous function} \cmark
    \end{itemize}
\end{enumerate}

\textbf{Topic 2: Structure and Training of Neural Networks}

\begin{enumerate}[resume]
    \item Why is it ineffective to merely combine multiple linear models in a sequential or stacked fashion?
    \begin{itemize}[label=$\square$]
        \item \textit{Multiplying matrices keeps the model linear} \cmark
        \item Sequential stacking of linear models creates a non-linear model by default.
        \item Optimization does not affect linearity
        \item \textit{Stacking linear models cannot capture non-linear interactions or complexities.} \cmark
    \end{itemize}

    \item What is the role of weights in a neural network?
    \begin{itemize}[label=$\square$]
        \item They adjust the learning rate
        \item \textit{They are parameters that help the network learn} \cmark
        \item \textit{They determine the strength of connections between neurons} \cmark
        \item They define the network’s architecture and layer structure
    \end{itemize}
\end{enumerate}

\textbf{Topic 3: Activation Functions}

\begin{enumerate}[resume]
    \item Which of the following are common activation functions in neural networks?
    \begin{itemize}[label=$\square$]
        \item Linear Function
        \item \textit{Sigmoid Function} \cmark
        \item Inverse Tangent Function
        \item \textit{Hyperbolic Tangent Function} \cmark
    \end{itemize}

    \item How does the ReLU activation function operate?
    \begin{itemize}[label=$\square$]
        \item \textit{Clamps negative values to zero} \cmark
        \item Normalizes all values to a range between -1 and 1
        \item \textit{Passes positive values unchanged}  \cmark
        \item Transforms negative values to positive values while leaving positive values unchanged
    \end{itemize}
\end{enumerate}

\subsubsection{Post-Test} \label{chap:post}
\textbf{Topic 1: Basics of Neural Networks}

\begin{enumerate}
    \item What is the primary function of a neural network?
    \begin{itemize}[label=$\square$]
        \item To organize data linearly
        \item \textit{To separate nonlinearities in data} \cmark
        \item \textit{To approximate complex functions} \cmark
        \item To execute matrix multiplication for feature extraction
    \end{itemize}

    \item Why are nonlinearities important in neural networks?
    \begin{itemize}[label=$\square$]
        \item \textit{They enhance the network’s ability to generalize} \cmark
        \item They reduce the computation time
        \item They normalize input data by scaling features
        \item \textit{They enable approximation of arbitrary non-linear functions } \cmark
    \end{itemize}
\end{enumerate}

\textbf{Topic 2: Structure and Training of Neural Networks}

\begin{enumerate}[resume]
    \item How is the dimensionality of weights (W) determined in a neural network?
    \begin{itemize}[label=$\square$]
        \item By the number of layers in the network
        \item \textit{By the number of neurons in the input and output layers} \cmark
        \item By the total count of biases across all layers
        \item \textit{By the product of neurons in adjacent layers} \cmark
    \end{itemize}

    \item How does a neuron in a neural network compute its output?
    \begin{itemize}[label=$\square$]
        \item By averaging the weighted inputs and using this average to adjust the learning rate
        \item By clustering inputs into groups and selecting outputs based on clusters
        \item \textit{By performing a dot product of its inputs and weights} \cmark
        \item \textit{By applying a nonlinearity to the input}  \cmark
    \end{itemize}
\end{enumerate}

\textbf{Topic 3: Activation Functions}

\begin{enumerate}[resume]
    \item Why are activation functions critical in neural networks?
    \begin{itemize}[label=$\square$]
        \item \textit{They enable the network to model intricate data distributions} \cmark
        \item They speed up training by accelerating convergence
        \item \textit{They introduce non-linearities, allowing the network to learn complex patterns} \cmark
        \item They normalize the data inputs for better performance
    \end{itemize}

    \item Which of the following is showing/describing the Sigmoid activation function?
    \begin{itemize}[label=$\square$]
        \item a) \begin{figure}[h]
	{\includegraphics[height=3.8cm, keepaspectratio]{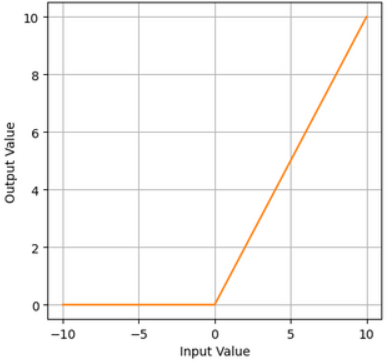}}
    \label{fig:relu}
\end{figure}
        \item b) $f(x) = max(0.1x,x)$
        \item c) \begin{figure}[t]
	{\includegraphics[height=3.8cm, keepaspectratio]{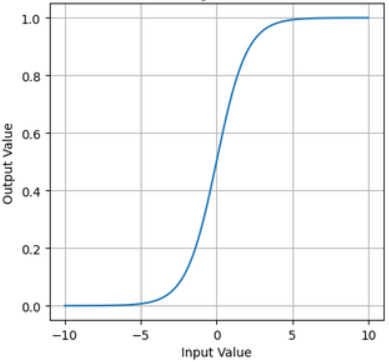}}
    \label{fig:sigmoid}
\end{figure}
 \cmark
 \pagebreak
        \item d) $f(x) = \dfrac{1}{1+e^{-x}} $  \cmark
    \end{itemize}
\end{enumerate}

\end{document}